\pdfoutput=1
\documentclass[11pt]{article}

\usepackage[]{emnlp2021}

\usepackage{times}
\usepackage{latexsym}
\usepackage{enumitem}
\usepackage{multirow}
\usepackage{blindtext}
\usepackage[T1]{fontenc}
\usepackage{tabu}

\usepackage[utf8]{inputenc}

\usepackage{microtype}
\usepackage{graphicx}
\usepackage{array}
\usepackage{booktabs}
\usepackage{adjustbox}
\usepackage{multirow}
\usepackage{cprotect, fancyvrb} %
\usepackage[normalem]{ulem}
\useunder{\uline}{\ul}{}

\title{HaRiM$^+$: Evaluating Summary Quality with Hallucination Risk}

\author{Seonil Son, \ Junsoo Park, \ Jeong-in Hwang, \ Junghwa Lee, \ Hyungjong Noh, \ Yeonsoo Lee \\
         NCSOFT NLP Center\\
         \texttt{\{deftson,junsoopark,jihwang,jleehhh0217,nohhj0209,yeonsoo\}@ncsoft.com}}

\begin{document}
\maketitle
\begin{abstract}
One of the challenges of developing a summarization model arises from the difficulty in measuring the factual inconsistency of the generated text.
In this study, we reinterpret the decoder overconfidence-regularizing objective suggested in \cite{miao-etal-2021-prevent} as a hallucination risk measurement to better estimate the quality of generated summaries.
We propose a reference-free metric, HaRiM${^+}$, which only requires an off-the-shelf summarization model to compute the hallucination risk based on token likelihoods.
Deploying it requires no additional training of models or ad-hoc modules, which usually need alignment to human judgments.
For summary-quality estimation, HaRiM$^+$ records state-of-the-art correlation to human judgment on three summary-quality annotation sets: FRANK, QAGS, and SummEval.
We hope that our work, which merits the use of summarization models, facilitates the progress of both automated evaluation and generation of summary.
\end{abstract}

\section{Introduction}
Although recent state-of-the-art summarization models have achieved remarkable performances \cite{lewis-etal-2020-bart, JMLR:v21:20-074, zhang2020pegasus}, appropriate metrics for measuring faithfulness of the generated summaries are still needed. The practice of measuring performance in the summarization task heavily relies on the N-gram matching based metric, ROUGE \cite{lin-2004-rouge}. 
Reportedly, ROUGE barely satisfies more than indicating lexical similarity \cite{maynez-etal-2020-faithfulness} and does not consider semantic dimensions of the generation, which current research needs of.

There have been numerous attempts to come up with faithfulness evaluation metrics \cite{novikova-etal-2017-need, peyrard-2019-studying}. Neural-based metrics have demonstrated good performances in estimating the factual consistency of a summary-article pair with semantic entailment \cite{kryscinski-etal-2020-evaluating, goyal-durrett-2020-evaluating}, question-answering framework \cite{wang-etal-2020-asking, scialom-etal-2021-questeval, scialom-etal-2019-answers}, and text generation \cite{NEURIPS2021_e4d2b6e6, xie-etal-2021-factual-consistency}. Most of the model-as-a-metric approach generally requires fine-tuning or complicated pipelines. Consequently, evaluating generated texts with recent model-as-a-metric methods has become cumbersome.\hfill \break
\indent With the increased demand for faithful generation models, it has come to a lot of attention on reformulating training objectives for purported for this \cite{zhang-etal-2022-conditional, liu-etal-2022-brio, holtzman-etal-2018-learning}. We focus on the training objective suggested in \cite{miao-etal-2021-prevent}, which directly targets hallucination problems in generating sentences given a source context. \citeauthor{miao-etal-2021-prevent} suggest that an overconfident decoder causes hallucination since the model excessively pays attention to the previously generated tokens over the source context which is in line with \cite{bowman-etal-2016-generating}.\hfill \break
\indent In this paper, we reinterpret the decoder overconfidence regularization term from \cite{miao-etal-2021-prevent} as \emph{hallucination risk} and recompose the objective to be practical for summary quality evaluation in various aspects.
Unlike other recent metrics \cite{NEURIPS2021_e4d2b6e6, xie-etal-2021-factual-consistency}, our metric, HaRiM${^+}$, detects hallucination in summary texts and evaluate their quality with the help of log-likelihood of summarization models. Also, HaRiM${^+}$ does not require complicated pipelines, further training, or modification of the generation model in use.\hfill \break
\indent We conduct experiments to verify the effectiveness of our metric on several summary quality estimation benchmarks. We test HaRiM${^+}$ on FRANK, annotation sets from QAGS, and SummEval, which provides multiple aspects of summary-quality judgements accompanied by summarization system outputs.   
Through quantitative and qualitative experiments, we demonstrate the robust performance of our metric HaRiM${^+}$, present the analysis of its inductive bias, and potential extension.

\section{Related Works} %

\subsection{Evaluation of Text Generation}
Automatic evaluation of generated text, despite its importance, has long relied on token-wise comparison against a reference target, and has been insufficient for reliably reflecting correctness and consistency. Most commonly used metrics, such as BLEU \cite{papineni-etal-2002-bleu}, ROUGE \cite{lin-2004-rouge}, and METEOR \cite{banerjee-lavie-2005-meteor}, are N-gram based metrics that compare token overlaps between candidate and reference texts. Model based metrics such as BERTScore \cite{zhang2019bertscore} use BERT representation of tokens, but such approaches have exhibited low correlation with human judgments of correctness for summarization datasets \cite{wang-etal-2020-asking}.

 As text generation models improve, sequence-to-sequence text generation models are increasingly being used for text quality evaluation.
BARTScore \cite{NEURIPS2021_e4d2b6e6} leverages the generation model's ability to assign higher probability to reference source-target pairs. 
PRISM \cite{thompson-post-2020-automatic} is a multilingual translation model that is used as a reference-to-candidate paraphraser.
COCO \cite{xie-etal-2021-factual-consistency} measures quality by estimating the effect of the language prior in text generation that contributes to hallucination.
The idea of using text generation models to estimate the log-likelihood of candidate sequences is conceptually simple yet has shown to be effective in evaluating text quality.
Our approach follows this line of research, but aims to improve the judgments of the consistency of the generated summary by adding a hallucination risk term.

\subsection{Hallucination Detection in Summarization}
Numerous works have addressed the need for an automatic way of detecting hallucination in generated summaries. This can be accomplished by reformulating detection problem into auxiliary tasks.
Textual entailment-based approaches consider the summary hallucination problem as a natural language inference (NLI) task, and leverage NLI classification models to score candidate summaries \cite{falke-etal-2019-ranking}.
QA-based approaches employ question generation and question answering models to generate questions from the candidate summary and to check the answerability of the question, respectively \cite{wang-etal-2020-asking, durmus-etal-2020-feqa, scialom-etal-2019-answers, scialom-etal-2021-questeval}.
\cite{goyal-durrett-2020-evaluating} propose to utilize dependency parser to classify whether each dependency arc is hallucinated.
QA-based approaches resemble the PYRAMID method \cite{nenkova-passonneau-2004-evaluating} and its automated descendants \cite{harnly2005automation,passonneau-etal-2013-automated,gao-etal-2019-automated} from a content selection perspective.

More direct approaches attempt to use models that are trained to distinguish artificially generated set of negative summaries. \citeauthor{kryscinski-etal-2020-evaluating} augments factual article-summary pairs to generate data for training a classification model. \citeauthor{zhou-etal-2021-detecting} employs a token-level prediction model to be trained on generated hallucination data.
All of the above methods require the generation of additional datasets and the training of auxiliary models.
In contrast, our approach only requires an off-the-shelf abstractive summarization model that needs no further training, and eliminates the need for preparing additional data.

\section{Method}
We describe the logic behind \emph{margin-based token-level objective} \cite{miao-etal-2021-prevent}, and reinterpret it as \emph{hallucination risk}. We then propose modifications to re-formulate the original objective to be feasible for evaluating text quality.

\subsection{\underline{Ha}llucination \underline{Ri}sk \underline{M}easurement (HaRiM)}
In encoder-decoder architectures, having the decoder relying too much on the decoder's context and less on the encoder's is a long known problem \cite{bowman-etal-2016-generating}.
\citeauthor{miao-etal-2021-prevent} introduced \emph{margin-based token-level objective} as a regularization term that prevents the decoder from focusing too much on the decoder-side context. 
Considering that hallucination refers to erroneous generation irrelevant to the source context, the regularization term can be reinterpreted as \emph{hallucination risk}. 
For source input text $X$ and target text $Y=\{y_0, y_1, ..., y_L\}$, the term HaRiM is defined as:
\begin{equation}
\label{eq1:hallucination_risk}
\mathrm{HaRiM} = \frac{1}{L}\sum_{i=0}^{L}(1 - p_{s2s})(1 - (p_{s2s} - p_{lm}) )
\end{equation}
where $p_{s2s}$ and $p_{lm}$ represent the token-likelihood of the sequence-to-sequence model (S2S) and that of the auxiliary language model (LM) respectively, and are defined as:
\begin{equation}
\label{eq2:notation}
p_{s2s}=p(y_i|y_{<i};X),~p_{lm} = q(y_i|y_{<i}) 
\end{equation} 

The S2S measures the probability of a target sequence with the knowledge of the encoder input $X$, while the LM does the same without $X$.
The value of HaRiM increases as the $p_{lm}$ overwhelms $p_{s2s}$. The value is weighted inversely by the S2S likelihood, thus maximizing when the S2S likelihood minimizes.

As described in the original paper, Equation \ref{eq1:hallucination_risk} is one of many ways of implementing the \emph{hallucination risk} using token likelihoods. However, after exploring many variations\footnote{Appendix Table  \ref{app:variations}}, we decide that the form in Equation \ref{eq1:hallucination_risk} works best for our purpose of quality estimation.

\subsection{Recomposing HaRiM for Feasible Evaluation}
\begin{figure}
\centering
\includegraphics[width=\columnwidth]{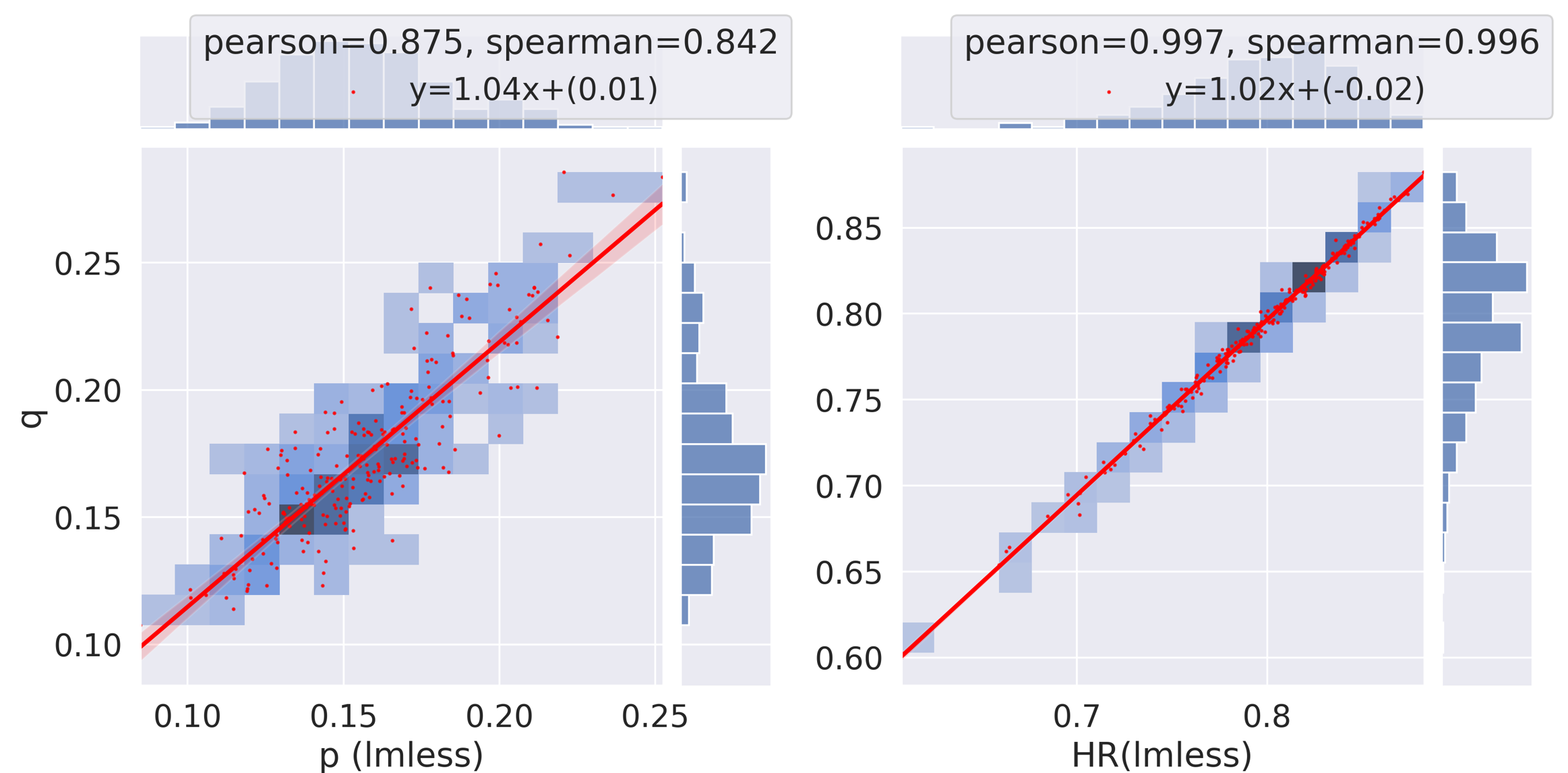}
\caption{Effects of replacing the auxiliary language model ($q(y_i|y_{<i})$) with an empty-sourced encoder-decoder model ($p(y_i|y_{<i};\{\}$). \textbf{Left} compares the values of $p_{lm}$, and \textbf{Right} compares the HaRiM values. The values are calculated on the summary-article pairs in FRANK benchmark. The high correlation of HaRiM suggests that the effect of replacement is minimal.}
\label{fig1:replacelm}
\end{figure}

\textbf{Replacing Auxiliary Language Model with Empty-Sourced Encoder-Decoder} \\
One of the challenges in applying hallucination risk to text evaluation is the requirement of the auxiliary language model ($q(\cdot)$ in Equation \ref{eq2:notation}) for the risk computation.
\citeauthor{miao-etal-2021-prevent} formulate the language model as an auxiliary decoder-only model that is jointly trained with the main encoder-decoder of the S2S model. 
However, when using an off-the-shelf summarization model for summary quality evaluation, this approach is infeasible because it needs a language model that should have been trained jointly with the summarization model, especially on a limited summarization dataset that can be insufficient for training a language model.
To avoid the joint training of language model, one can consider using a pre-trained language model to replace the auxiliary model.
However this approach is also infeasible because the tokenization and vocabulary of the language model must match the ones of the S2S model.

Instead we consider re-purposing the entire encoder-decoder from the summarization model itself as a language model.
In this way, the LM model is simply the S2S model itself, but works as an LM when it receives an empty source text (denoted as $\{\}$) as the encoder input.
This eliminates the need for an additional model, and automatically solves the tokenization and vocabulary issue as well.
Thus we replace the  $p_{lm}$ from auxiliary language model likelihood ($q(\cdot)$) to empty-sourced S2S likelihood as the following\footnote{We implemented empty input (\{\}) as a sequence with only begin and end of the sequence token, namely [BOS], and [EOS]}:
\begin{equation}
\label{eq3:notation}
p_{lm} = p_{s2s}(y_i|y_{<i};\{\})
\end{equation} %
We test the validity of such modified use of S2S model as the LM model when calculating the hallucination risk.
We compare the hallucination risk value when replacing $p_{lm}$ from auxiliary language model to empty-sourced S2S.
The results in Figure \ref{fig1:replacelm} show that hallucination risk HaRiM calculated with empty-sourced S2S is almost perfectly linear with the counterpart computed with the auxiliary model ($\rho=.997$), thus $p_{lm}$ is replaceable as the Equation \ref{eq3:notation} in computation of HaRiM.\footnote{$p_{lm}$ is not negligible for computing HaRiM (Appendix, Figure \ref{fig:HaRiMvVars}).}

\noindent\textbf{Accompanying HaRiM with Log-likelihood (HaRiM${^+}$)} \\ 
A broad range of factors for text quality estimation makes evaluation task hard because it varies according to the generation task.
An implicit way of measuring overall generation quality is to use token likelihood of high-performing text-generation models as reported in \cite{NEURIPS2021_e4d2b6e6}. %
We find that accompanying sequence-to-sequence log-likelihood ($\mathrm{log}p_{s2s}$) of tokens to hallucination risk helps estimating comprehensive quality more than factual consistency, such as fluency.
As in Equation \ref{eq3:ce_7hr}, hallucination risk is scaled with a hyperparameter $\lambda$, and the log-likelihood of tokens is added to form $\mathrm{HaRiM^+}$.
\begin{equation}
\label{eq3:ce_7hr}
\mathrm{HaRiM^+} = \frac{1}{L}\sum_i^L{ \mathrm{log}(p(y_i|y_{<i};X))}-\lambda*\mathrm{HaRiM}
\end{equation} 
In our experiments, we used $\lambda=7$, which is a value coherent with the works of \citeauthor{miao-etal-2021-prevent}.\footnote{$\lambda$ is determined primarily based on metric correlation to human judgements, but with the consideration of scales of each (Appendix, Figure \ref{fig:scorescales}).}

\section{Experiments}
\subsection{Summarization Quality Benchmarks}
\subsubsection{Factual Consistency Benchmarks}
We choose FRANK \cite{pagnoni-etal-2021-understanding}, and QAGS annotations \cite{wang-etal-2020-asking} as benchmarks for assessing the metrics' power to resolve the factuality of article-summary pairs.
FRANK and QAGS contain 2246 and 470 pairs, respectively, of article and system-generated summary from CNN-DailyMail \cite{nallapati-etal-2016-abstractive}, as well as BBC-XSUM \cite{narayan-etal-2018-dont} corpora.
Every pair in the benchmark contains human judgement on factuality.
Both benchmarks have similar purpose and annotation format, but differ in annotating environment and aggregation process of the annotations. For FRANK, factual pairs are the intact examples remaining after the annotating errors of each summary introduced by number of annotators, but in QAGS, annotators are directly asked to label each pair if it is factually consistent.
We report separate results on each testbed.

In the case of FRANK, the authors recommend measuring partial correlation by considering the confounding variable, the summarization system where summaries are generated from, which can undermine the gaps between metric performances. However, we do not follow this suggestion and conduct experiments with the same setting as others.\footnote{We provide a graphical model representing our claim in Appendix (Figure \ref{fig:graphicalmodel}). Reporting partial correlation to consider the bias introduced by generation system artifacts in the text might help alleviate the vulnerability of a metric, but, in principle, metric does not refer to any other attribute than the text. Thus we decided not to follow the practice of the original benchmark.}

\subsubsection{Comprehensive Quality Benchmark}
SummEval \cite{fabbri-etal-2021-summeval} contains 1600 annotated article-generated summary pairs from 16 summarization systems.
The benchmark lets annotators answer about four criteria that a good summary pair should satisfy: coherence, consistency, fluency, and relevance. Each criterion attributes to whether a certain summary is well-organized in structure, factually consistent, grammatically fluent, and containing relevant information regarding the message of the article, respectively.
SummEval is comprised of outputs from both abstractive and extractive summarization models which allows dimensional analysis for metrics' performance.
We use only the annotations from experts, excluding the ones from turkers, in accordance with the other works' practice using the SummEval for benchmarking \cite{scialom-etal-2019-answers, scialom-etal-2021-questeval, liu2022maskeval}.\footnote{In Appendix  Figure \ref{app:exponly}, We also discuss about reasons why turker annotations are less preferred in discussion section, which supports the arguments from the original authors.}

\subsection{Measures for Meta-evaluation of Metrics} 
Measures for describing correlation between two variables are as follows: 
\begin{itemize} %
  \itemsep0em
  \item \textbf{Kendall's $\tau$} measures how good the metric is ranking the examples (article-summary pairs) in order of human judgement. 
  \item \textbf{Spearman's $r$} assesses how well the relation between the metric and human judgement can be described as monotonic function. 
  \item \textbf{Pearson's $\rho$} measures how linear the metric score is. This may not represent monotonic increment or decrement to the human annotations, but represents proper scaling of the metric; i.e. A metric score should increase linearly according to increment of the judgement score.
\end{itemize}
All three coefficients range from 0 (independent) to 1 (completely correlated). 
We report metric-human correlation in $\tau$, and metric-metric correlation with $\rho$. 
We find that trends of all three measures move together in our case, and we report $\tau$ correlation as the primary measure in our meta-evaluation results in Table \ref{tab:qmain}.
Correlations in other measures are reported on Appendix (Table \ref{tab:spearman_pearson}) for further information.

\subsection{Metrics}
\subsubsection{Traditional Metrics}
We benchmark traditional N-gram matching baselines; ROUGE-1, 2, L \cite{lin-2004-rouge}, METEOR \cite{banerjee-lavie-2005-meteor}, sacreBLEU \cite{riddell-etal-2021-call} on three benchmarks.\footnote{For implementation details, please refer to Appendix \ref{app:implementation_details}.}
For matching-based metrics, we test not only matching to the reference summaries but also to the article (noted as `\_art'), which is reported to benefit metrics assessing factual coverage of the summary \cite{pagnoni-etal-2021-understanding}.
Additionally, we report some of the relevant statistics; length, and ratio of novel N-gram \cite{fabbri-etal-2021-summeval} in the summary as a metric to compare.

\subsubsection{Unsupervised Matching}
We also test our metric against the relatively recent matching-based metric based on contextual embedding, BERTScore \cite{zhang2019bertscore}.
BERTScore borrows representation power of the pretrained masked language model, BERT \cite{devlin-etal-2019-bert}, to match contextualized embeddings of two texts.  

We used roberta-large \cite{DBLP:journals/corr/abs-1907-11692} checkpoint provided as default by the package.\footnote{
    \url{https://github.com/Tiiiger/bert_score}
}
As done for N-gram metrics, matching toward article is also reported with `\_art' notation.

\subsubsection{Text Generation Task as an Evaluation}
BARTScore \cite{NEURIPS2021_e4d2b6e6} reformulated text quality evaluation as a text generation problem. 
BARTScore depends on the log-likelihood of the fine-tuned BART model to score the quality of the text; averaged log-likelihood of a text is a quality estimation.
In our experiments, we test two versions of BARTScore introduced in the original paper. 
One is BART-large fine-tuned on CNN-DailyMail corpus \cite{lewis-etal-2020-bart}, the other is further fine-tuned to ParaBank2 corpus \cite{hu-etal-2019-large} to better capture factual consistency of the article-summary pairs.\footnote{
        Model checkpoints for BARTScore are from \url{https://huggingface.co/facebook/bart-large},\url{https://github.com/neulab/BARTScore}.
}
We also augment BARTScore with hallucination risk to test its correlation toward human judgements.
Another objective used as a metric is from CBMI \cite{zhang-etal-2022-conditional}, which re-weights negative log-likelihood loss with the conditional bilingual mutual information approximated from token statistics. 
We flipped the sign of the loss for it to work as higher-better metric.\footnote{For detailed information of implementation, refer to Appendix \ref{app:cbmi}.}

\subsubsection{Question Answering as an Evaluation}
Metrics in QA generally require question generation and answering modules that check whether the summary is factually supported by the article.
We refer to FEQA \cite{durmus-etal-2020-feqa} and QAGS \cite{wang-etal-2020-asking} to examine the performance of the QA-based metrics.   
We benchmark QAGS on two factuality benchmarks, FRANK and QAGS.
On QAGS annotations, we re-run QAGS from the original repository\footnote{\url{https://github.com/W4ngatang/qags}} to score towards the benchmark. 
On FRANK, we reused the QAGS and FEQA scores publicly shared on FRANK repository.\footnote{\url{https://github.com/artidoro/frank}}

\subsubsection{Proposed Method: HaRiM${^+}$}
HaRiM${^+}$, our proposed method, exploits summarization model for calculating HaRiM and complement it with log-likelihood, as in Equation \ref{eq3:ce_7hr} to make the final metric score.
We use the same summarization model checkpoints as BARTScore as described above for direct comparison: BART-large+cnn \cite{lewis-etal-2020-bart}, and BART-large+cnn+para \cite{NEURIPS2021_e4d2b6e6}.
In the ablation study (Section \ref{subsec:ablation_study}), we added another checkpoint, BRIO \cite{liu-etal-2022-brio} which also has the same architecture with BART-large.

\section{Results}
In the followings, we report (1) metric to human judgement correlation in Kendall's $\tau$ rank coefficient, 
and (2) qualitative examples that reveals inductive bias of the hallucination risk (HaRiM${^+}$) we proposed.
Comparisons with reported values of several other works are attached to Appendix (Table \ref{tab:reported_comp}). 

\subsection{Metric-Human Correlation}
Table \ref{tab:qmain} shows the metric to human judgement (segment-level)\footnote{system level correlation reported in Table \ref{tab:app_syslv_corr}} correlation.
Proposed HaRiM${^+}$ records highest Kendall's $\tau$ in most criteria of \emph{CNN/DailyMail} based benchmarks.
To thoroughly show the significance test result, we attach permutation test matrix on Figure \ref{fig:FRANK_CNN_permutation} in Appendix.
Because HaRiM${^+}$ and BARTScore shares the same summarization model, both metrics with respective models show similar scoring patterns. 
HaRiM${^+}$ records mostly highest correlation toward human judgements except several settings (XSUM, and SummEval-Relevance).
For SummEval relevance score benchmark, BERTScore P\_art outperforms the HaRiM${^+}$ (BART\_large + cnn) by 0.024 points, which indicates BERTScore P\_art is 1.2\%p better at ranking hallucinated results.
In FRANK-XSUM benchmark, despite using a summarization model trained on \textit{CNN/DailyMail}, HaRiM${^+}$ records high score ($\tau=0.141$ compared to $\tau=0.151$ of BERTScore P).
On FRANK-CNNDM, we perform a permutation test to confirm that HaRiM${^+}$ outperforms the others with the confidence $>$.95 which is attached to the Appendix (Table \ref{fig:metmet_corr_pearson_cnn}) for space issue.\footnote{
    Several notable observations in metric-metric correlation had to be pushed back to Appendix (e.g. NovelNgram highly correlates (>.6) to BERTScore\_art, and HaRiM${^+}$, but HaRiM${^+}$ and BERTScore\_art are not).
}
Overall, HaRiM${^+}$ records robust performances in ranking the summary pairs according to the human judgement for CNN-DailyMail corpus examples which the core model is trained to, while it also scored high on XSUM corpus.

\begin{table*}[ht]
\centering
\resizebox{\textwidth}{!}{
\tabulinesep=0.8mm
\begin{tabu}{lcccccccc}
\multicolumn{1}{l|}{}                                        & \multicolumn{6}{c|}{\textbf{CNNDM}}                                                                                                & \multicolumn{2}{c}{\textbf{XSUM}}         \\
\multicolumn{1}{l|}{Kendall's $\tau$}                                        & \textbf{FRANK}      & \textbf{QAGS}       & \multicolumn{4}{c|}{\textbf{SummEval}}                                                 & \textbf{FRANK}      & \textbf{QAGS}       \\ \hline
\multicolumn{1}{l|}{\textbf{Metrics}}                        & \textbf{Factuality} & \textbf{Factuality} & \textbf{Con}   & \textbf{Coh}   & \textbf{Flu}   & \multicolumn{1}{c|}{\textbf{Rel}}   & \textbf{Factuality} & \textbf{Factuality} \\ \hline
\textbf{N-gram-matching}                                     &                     &                     &                &                &                &                                     &                     &                     \\ \hline
\multicolumn{1}{l|}{ROUGE 1}   & 0.182               & -0.052              & 0.105          & 0.123          & 0.062          & \multicolumn{1}{c|}{0.209}          & 0.125               & 0.110               \\
\multicolumn{1}{l|}{ROUGE 2}                                 & 0.135               & -0.107              & 0.101          & 0.097          & 0.048          & \multicolumn{1}{c|}{0.153}          & 0.128               & 0.097               \\
\multicolumn{1}{l|}{ROUGE L}                                 & 0.141               & -0.072              & 0.091          & 0.113          & 0.061          & \multicolumn{1}{c|}{0.164}          & 0.117               & 0.090               \\
\multicolumn{1}{l|}{METEOR}                                  & 0.198               & 0.053               & 0.125          & 0.116          & 0.070          & \multicolumn{1}{c|}{0.223}          & 0.121               & 0.115               \\
\multicolumn{1}{l|}{sacreBLEU}                               & 0.136               & -0.085              & 0.080          & 0.167          & 0.088          & \multicolumn{1}{c|}{0.131}          & 0.113               & 0.012               \\ \hline
\multicolumn{1}{l|}{ROUGE 1\_art}                            & 0.185               & 0.243               & 0.111          & 0.036          & 0.058          & \multicolumn{1}{c|}{0.127}          & -0.003              & -0.074              \\
\multicolumn{1}{l|}{ROUGE 2\_art}                            & 0.249               & 0.315               & 0.195          & 0.072          & 0.119          & \multicolumn{1}{c|}{0.165}          & 0.027               & 0.069               \\
\multicolumn{1}{l|}{ROUGE L\_art}                            & 0.225               & 0.305               & 0.203          & 0.097          & 0.123          & \multicolumn{1}{c|}{0.050}          & 0.010               & -0.019              \\
\multicolumn{1}{l|}{METEOR\_art}                             & 0.174               & 0.234               & 0.112          & 0.009          & 0.071          & \multicolumn{1}{c|}{0.091}          & 0.004               & -0.052              \\
\multicolumn{1}{l|}{sacreBLEU\_art}                          & 0.153               & 0.245               & 0.091          & 0.042          & 0.035          & \multicolumn{1}{c|}{}               & -0.038              & -0.139              \\ \hline
\textbf{N-gram stats}                                        &                     &                     &                &                &                &                                     &                     &                     \\ \hline
\multicolumn{1}{l|}{NovelNgram\_4}                           & 0.275               & {\ul 0.392}               & 0.221          & 0.203          & 0.173          & \multicolumn{1}{c|}{0.205}          & 0.017               & 0.056               \\
\multicolumn{1}{l|}{NovelNgram\_3}                           & 0.273               & 0.370               & 0.218          & 0.208          & 0.171          & \multicolumn{1}{c|}{0.208}          & 0.064               & 0.080               \\
\multicolumn{1}{l|}{NovelNgram\_2}                           & 0.259               & 0.327               & 0.199          & 0.209          & 0.150          & \multicolumn{1}{c|}{0.207}          & 0.053               & {\ul 0.129}         \\
\multicolumn{1}{l|}{NovelNgram\_1}                           & 0.219               & 0.201               & 0.090          & 0.190          & 0.068          & \multicolumn{1}{c|}{0.173}          & 0.091               & 0.120               \\
\multicolumn{1}{l|}{Length (no. tokens)}                     & 0.187               & 0.185               & 0.078          & 0.033          & 0.000          & \multicolumn{1}{c|}{0.000}          & -0.111              & -0.132              \\ \hline
\textbf{Contextual Embedding}                                &                     &                     &                &                &                &                                     &                     &                     \\ \hline
\multicolumn{1}{l|}{BERTScore P}                             & 0.168               & -0.067              & 0.041          & 0.229          & 0.097          & \multicolumn{1}{c|}{0.192}          & \textbf{0.151}      & 0.016               \\
\multicolumn{1}{l|}{BERTScore R}                             & 0.250               & 0.017               & 0.125          & 0.241          & 0.097          & \multicolumn{1}{c|}{0.299}          & 0.107               & 0.058               \\
\multicolumn{1}{l|}{BERTScore F1}                            & 0.232               & -0.029              & 0.079          & 0.267          & 0.111          & \multicolumn{1}{c|}{0.267}          & 0.142               & 0.036               \\ \hline
\multicolumn{1}{l|}{BERTScore P\_art}                        & 0.301               & 0.331               & {\ul 0.266}    & {\ul 0.308}    & {\ul 0.236}    & \multicolumn{1}{c|}{\textbf{0.308}}    & 0.038               & -0.039              \\
\multicolumn{1}{l|}{BERTScore R\_art}                        & 0.360               & 0.365               & 0.141          & 0.153          & 0.112          & \multicolumn{1}{c|}{0.234}          & 0.144               & -0.022              \\
\multicolumn{1}{l|}{BERTScore F1\_art}                       & 0.358               & 0.365         & 0.230          & 0.256          & 0.192          & \multicolumn{1}{c|}{0.307}          & 0.111               & -0.040              \\ \hline
\textbf{Neural entailment}                                   &                     &                     &                &                &                &                                     &                     &                     \\ \hline
\multicolumn{1}{l|}{FactCC \cite{kryscinski-etal-2020-evaluating}}                                  &  0.376         &                     &                &                &                & \multicolumn{1}{c|}{}               & 0.071               &                     \\
\multicolumn{1}{l|}{Dep Entail \cite{goyal-durrett-2020-evaluating}}                              & 0.342               &                     &                &                &                & \multicolumn{1}{c|}{}               & 0.092               &                     \\ \hline
\textbf{Q\&A based}                                          &                     &                     &                &                &                &                                     &                     &                     \\ \hline
\multicolumn{1}{l|}{FEQA \cite{durmus-etal-2020-feqa}}                                    & -0.008              &                     &                &                &                & \multicolumn{1}{c|}{}               & 0.006               &                     \\
\multicolumn{1}{l|}{QAGS \cite{wang-etal-2020-asking}}                                    & 0.206               & 0.274               &                &                &                & \multicolumn{1}{c|}{}               & -0.006              & \textbf{0.153}      \\ %
\multicolumn{1}{l|}{QAEval-F1 \cite{deutsch-etal-2021-towards}}                                    &                &                &                &                &                & \multicolumn{1}{c|}{0.220*}               & -0.006              & \textbf{0.153}      \\\hline 
\textbf{Text Generation based}                               &                     &                     &                &                &                &                                     &                     &                     \\ \hline
\multicolumn{1}{l|}{CBMI (BART\_base + cnn)}                & 0.058               & 0.026               & 0.152          & -0.029         & 0.023          & \multicolumn{1}{c|}{0.208}          & -0.077              & -0.041              \\
\multicolumn{1}{l|}{BARTScore (BART\_large+cnn) \cite{NEURIPS2021_e4d2b6e6}}          & 0.413               & 0.470               & 0.197          & 0.310          & 0.181          & \multicolumn{1}{c|}{0.263}          & 0.137               & 0.072               \\
\multicolumn{1}{l|}{BARTScore (BART\_large+cnn+para) \cite{NEURIPS2021_e4d2b6e6}} & {\ul 0.392}               & 0.416               & 0.259          & 0.301          & 0.238          & \multicolumn{1}{c|}{0.278}          & {\ul 0.145}         & 0.031               \\ \hline
\textbf{Proposed}                                            &                     &                     &                &                &                &                                     &                     &                     \\ \hline
\multicolumn{1}{l|}{\textbf{HaRiM${^+}$ (BART\_large + cnn)}}         & \textbf{0.424}      & \textbf{0.478}      & 0.251          & \textbf{0.315} & 0.210          & \multicolumn{1}{c|}{{\ul 0.284}} & 0.136               & 0.076               \\
\multicolumn{1}{l|}{HaRiM${^+}$ (BART\_large + cnn + para)}            & 0.399               & 0.401               & \textbf{0.281} & 0.293          & \textbf{0.245} & \multicolumn{1}{c|}{0.282}          & 0.141               & 0.028               \\ \hline
\end{tabu}}
\caption{\label{tab:qmain}Metric-to-human judgement correlation (segment level) reported in Kendall's $\tau$. \textbf{Bold}-face values are the largest correlating metrics, underlined are second-large values amongst the metrics. HaRiM${^+}$ outperforms others in most criteria. SummEval's quality criteria; consistency, coherence, fluency, and relevance are abbreviated as Con, Coh, Flu, and Rel respectively. We provide permutation test result and results in Spearman's $r$ and Pearson's $\rho$ in Appendix (Figure \ref{fig:FRANK_CNN_permutation}, Table \ref{tab:spearman_pearson}). In Table \ref{tab:reported_comp}, we also provide comparisons to reported values that could not be directly presented above. \textcolor{gray}{*:correlation value taken from \cite{deutsch-etal-2021-towards}}} %
\end{table*}

\subsection{Ablation Study: Effect of Accompanying Log-likelihood}
\label{subsec:ablation_study}
We conduct ablation study on HaRiM${^+}$ varying the model checkpoints. 
HaRiM${^+}$ is compared to each term used in single: log-likelihood, and the regularization term only (HaRiM).
Table \ref{tab:my-table} shows the results for the average scores across all four SummEval criteria; the table indicates that accompanied use of log-likelihood with HaRiM (that is, HaRiM$^+$) helped complementing the metric performance.
\begin{table}[]
\resizebox{\columnwidth}{!}{%
\begin{tabular}{l|ccc}
\hline
Checkpoints & Log-likelihood & HaRiM & HaRiM${^+}$     \\ \hline
\textbf{BART-large + cnn}                  & 0.238          & \textbf{0.279} & 0.265          \\
\textbf{BART-large + cnn + para}           & 0.269          & 0.256          & \textbf{0.275} \\
\textbf{BRIO \cite{liu-etal-2022-brio}}                 & 0.262          & 0.252          & \textbf{0.265} \\ \hline
\end{tabular}}
\caption{Effect of accompanied use of log-likelihood and regularization term HaRiM (HaRiM $\rightarrow$ HaRiM$^+$). The values are average of four $\tau$ correlation from SummEval.}
\label{tab:my-table}
\end{table}

\subsection{Qualitative Analysis: Detecting Hallucinations}
\begin{table*}[ht]
\centering
\resizebox{\textwidth}{!}{
\tabulinesep=0.6mm
\begin{tabu}{@{}clcc@{}}
\toprule
\multicolumn{4}{c}{\textbf{Source Article}} \\ \midrule
\multicolumn{4}{m{22.6cm}}{The view that Manchester City’s chance at defending their Premier League title has been ruined through bad spending gathered pace after they were defeated by a club whose entire team cost less than half one of their substitutes. Crystal Palace’s XI on Monday night may only have been worth a mere £17m, but left back Martin Kelly still made it through a City defence deemed good enough to keep £40m signing Eliaquim Mangala on the bench to tee up a chance for Wilfried Zaha just 60 seconds into the game. Mangala joined from Porto in August last year and is contracted to City until June 2019. Eliaquim Mangala (green bib) prepares to come on but he never made it off the Manchester City bench However, striker Glenn Murray succeeded in putting another dent in City’s chances of redeeming themselves after a run of four losses away, when he scored Palace’s first goal. Murray cost Palace nothing when joined from arch rivals Brighton in 2011. Jason Puncheon, signed for a comparative pittance of £1.9m, delivered City their final blow with a goal from a finely executed free-kick. Glenn Murray (left) cost Palace nothing four years ago yet found a way past the City defence Another expensive City player, £24m-man Yaya Toure, got his team back in the game with 12 minutes left, but they couldn’t penetrate Palace’s defence to find an equaliser and a 2-1 defeat leaves them nine points adrift of the top. Toure joined from Barcelona in July 2010 and is contracted to City until 2017. After spending a total of £500m pounds on transfer fees, City might have expected to be higher than a precarious fourth in the league, but judging by their latest results, it’s teams like Crystal Palace that seem to be getting their value for money. Mangala has endured a miserable first season at the Etihad Stadium since his £40million move} \\ \midrule
\multicolumn{1}{c|}{\textbf{Model}} & \multicolumn{1}{c|}{\textbf{Summary}} & \multicolumn{1}{c|}{\textbf{$\mathrm{HaRiM}^+$ Score $\uparrow$}} & \textbf{Score Gain $\uparrow$} \\ \midrule
\multicolumn{1}{c|}{\textbf{Reference}} & \multicolumn{1}{m{13.2cm}|}{manchester city beaten 2-1 by crystal palace on easter monday . \# 40m signing eliaquim mangala was left on the bench . crystal palace 's entire starting xi cost just \# 17million . click here for all the latest manchester city news .} 
& \multicolumn{1}{c|}{1.7218} 
& - \\ \midrule

\multicolumn{1}{c|}{\textbf{\begin{tabular}[c]{@{}c@{}}Self-generation\\ (BART-large+cnn)\end{tabular}}} & \multicolumn{1}{m{13.2cm}|}{manchester city lost 2-1 to crystal palace at the etihad on monday night. crystal palace's entire team cost less than half one of manchester city's substitutes. eliaquim mangala and yaya toure were both left on the bench. city have spent a total of £500m on transfer fees so far this season.} 
& \multicolumn{1}{c|}{3.7006} 
& +1.9788 \\ \midrule

\multicolumn{1}{c|}{\textbf{\begin{tabular}[c]{@{}c@{}}BottomUpSummary\\ (Factuality=\textcolor{red}{0.0})\end{tabular}}} & \multicolumn{1}{m{13.2cm}|}{crystal palace 's xi is contracted to city until june 2019 . jason puncheon signed for \# 1.9 m from porto in august last year . glenn murray has scored four goals in the premier league .} 
& \multicolumn{1}{c|}{1.0265} 
& -0.6953  \\ \midrule

\multicolumn{1}{c|}{\textbf{\begin{tabular}[c]{@{}c@{}}Reference\\ (w/ wrong subject)\end{tabular}}} & \multicolumn{1}{m{13.2cm}|}{manchester city beaten 2-1 by crystal palace on easter monday . \# 40m signing \textcolor{red}{wilfried zaha} was left on the bench . crystal palace 's entire starting xi cost just \# 17million . click here for all the latest manchester city news .} 
& \multicolumn{1}{c|}{1.5571} 
& -0.1647 \\ \midrule

\multicolumn{1}{c|}{\textbf{\begin{tabular}[c]{@{}c@{}}Reference\\ (w/ negation)\end{tabular}}} & \multicolumn{1}{m{13.2cm}|}{manchester city beaten 2-1 by crystal palace on easter monday . \# 40m signing eliaquim mangala \textcolor{red}{was not} left on the bench . crystal palace 's entire starting xi cost just \# 17million . click here for all the latest manchester city news .} 
& \multicolumn{1}{c|}{1.5298} 
& -0.192 \\ \bottomrule
\end{tabu}}
\caption{\label{tab:qual_neg1_}Testing $\mathrm{HaRiM}^+$ metric under hallucination detecting scenario. The words highlighted red are hallucinated information deliberately injected to the reference. BottomUpSummary refers to abstractive summarization system suggested in \cite{gehrmann-etal-2018-bottom}.}
\end{table*}

We test the HaRiM${^+}$ (BART-large+cnn) under hallucination detecting scenario to provide hint for how HaRiM${^+}$ behaves in various summary outputs. 
In Table \ref{tab:qual_neg1_}, we randomly pick an article from \textit{CNNDailyMail} split of the FRANK benchmark and prepare several summaries.
We collected the following five summaries to pair with the article: (1) reference target summary, (2) summary generated from BART-large+cnn (Self-generation), (3) unfactual summary of summarization model (displayed example is generated by BottomUpSummary \cite{gehrmann-etal-2018-bottom}), (4) reference summary permutation with wrong subject, which contains wrongly-injected subject entity from the source article, and (5) a negated reference summary.

\noindent As shown in Table \ref{tab:qual_neg1_}, we align the summary with HaRiM${^+}$ (BART-large+cnn) score and its score gain compared to the reference summary score. 
HaRiM${^+}$ metric ranks the summaries in order of self-generated$>$reference$>$permuted references$>$wrong generation.
We attribute the HaRiM${^+}$ metric's preference toward self-generation to inductive bias: both the self-generation model and HaRiM${^+}$ evaluation model are the exact twins. 
To roughly put, the self-generation model works as an oracle summary generator for the metric. 
The inductive bias of HaRiM${^+}$ metric will be discussed further with quantitative evidence in Section \ref{inductivebias}.
The trend of ranking factual human-written summaries over unfactual summaries, which includes permutated references, are observed constantly throughout the \textit{CNNDailyMail} corpus examples. 
We provide several more examples in Appendix (Table \ref{tab:qual_neg1}, \ref{tab:quala}, \ref{tab:qual_neg2}, \ref{tab:qual_pos1}, and \ref{tab:qual_pos2}).

\section{Discussion}
\subsection{Inductive Bias} \label{inductivebias}
\begin{figure}
\centering
\includegraphics[width=\columnwidth]{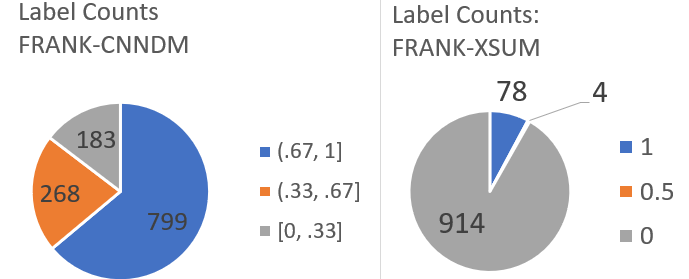}
\caption{\label{fig:frank_xsum_dist}Factuality label counts from FRANK benchmark. Legend shows the value of factuality annotation, varying from 0 (unfactual) to 1 (factual). The  factuality labels for XSUM corpus are almost binary.}
\end{figure}

As mentioned in qualitative analysis, the metric has inductive bias of preferences toward summaries generated by abstractive summarization systems.
Proposed HaRiM${^+}$ prefers self-generated summary (i.e. summary generated by the same summarization model the scorer depending on) to human written references.
Another hint for this bias could be found when we dissect the SummEval benchmark results into abstractive and extractive summary splits.
In Table \ref{tab:abs_prefer}, not only log-likelihood but also regularization term, HaRiM, both prefer outputs from abstractive system.
As summary text becomes similar to the evaluating summarization model's likely output, generation-based metrics (including HaRiM${^+}$) become more generous at scoring.
In other word, how bad the assessed summary would not be a problem if the summarizer used for evaluation resembles the system which wrote the summary being assessed.
In this context, using the model trained on too noisy dataset, without proper regularization would result in unreliable evaluation.
Figure \ref{fig:frank_xsum_dist} shows how noisy summarization models could be trained under-regularized; most of the output summary trained on \textit{XSUM} with MLE strategy contain errors.
Therefore, we decide not to exploit summarization model fine-tuned on \emph{XSUM} even if it could result in better correlation on FRANK/QAGS-XSUM splits.

\begin{table}[h]
\centering
\resizebox{\columnwidth}{!}{
\begin{tabular}{@{}llccc@{}}
\toprule
                                                                                &                & \textbf{abstractive} & \textbf{extractive} & \textbf{$\Delta$} \\ \hline
\multirow{3}{*}{\textbf{\begin{tabular}[c]{@{}l@{}}BART\\-Large\end{tabular}}} & log-likelihood & 0.266        & 0.160        & 0.106     \\
                                                                                & HaRiM             & 0.303        & 0.174        & 0.129     \\
                                                                                & HaRiM${^+}$             & 0.293        & 0.168        & 0.125     \\ \hline
\multirow{3}{*}{\textbf{BRIO}}                                                  & log-likelihood & 0.308        & 0.143        & 0.165     \\
                                                                                & HaRiM              & 0.295        & 0.117        & 0.177     \\
                                                                                & HaRiM${^+}$             & 0.311        & 0.137        & 0.174     \\ \hline
\multirow{3}{*}{\textbf{\begin{tabular}[c]{@{}l@{}}BART\\ -Score\end{tabular}}} & log-likelihood & 0.296        & 0.168        & 0.128     \\
                                                                                & HaRiM             & 0.280        & 0.150        & 0.130     \\
                                                                                & HaRiM${^+}$             & 0.303        & 0.166        & 0.137     \\ \hline
\textbf{Average}                                                                    &                & 0.295        & 0.154        & 0.141     \\ \bottomrule
\end{tabular}
}
\caption{\label{tab:abs_prefer} Averaged $\tau$ correlation on SummEval. $\Delta$ indicates difference of $\tau$ coefficients measured toward abstractive and extractive summaries.}
\end{table}

\subsection{Metric Performance of HaRiM${^+}$ in Machine Translation}
We also tested our metric, HaRiM${^+}$, on WMT20 metrics task \cite{mathur-etal-2020-results} to see whether HaRiM${^+}$ works in the machine translation domain (Table \ref{tab:mt}).
WMT20 DA annotation contains machine translation pairs of language pairs accompanied with human judgements of quality. 
We find that there is little improvement in correlation to human annotation in several language pairs, but it is not significant in average of all language pairs. 
In case of WMT20 metrics task, performance of the generation-based metrics seems to rely heavily on generation model checkpoints and its train corpus distribution rather than the hallucination risk consideration. 
As WMT metrics task has a broad range of  dimensions to explore, we leave this as a future remark for generation-based evaluation metrics and text generation models.

\begin{table}[]
\resizebox{\columnwidth}{!}{%
\begin{tabular}{lcccc}
\hline
                     & \multicolumn{2}{c}{\textbf{sys($\rho$)}} & \multicolumn{2}{c}{\textbf{seg($\tau$)*}} \\ \cline{2-5} 
                     & \textbf{all}  & \textbf{all-out} & \textbf{all}  & \textbf{all-out} \\ \hline
\multicolumn{5}{l}{\textbf{(1) BART-large+cnn+para$\rightarrow$MBART50\_m2m}} \\ \hline
Log-likelihood       & -0.001        & -0.005           & -0.020        & -0.024           \\
HaRiM${^+}$                   & 0.002         & 0.000            & -0.016        & -0.020           \\ \hline
\multicolumn{5}{l}{\textbf{(2) Log-likelihood$\rightarrow$HaRiM${^+}$}}                                 \\ \hline
BART-large+cnn+para  & +0.001        & 0                & 0             & -0.001           \\
PRISM(m39v1)         & 0             & 0                & 0             & +0.001           \\
MBART50\_m2m         & 0             & +0.002           & +0.001        & +0.002           \\ \hline
\end{tabular}%
}
\caption{\label{tab:mt}Change of generation-based metric performance according to (1) model weight change (2) applying HaRiM${^+}$. All results are averaged over language pairs from data supported by each model (i.e. BART-large+cnn+para averages the results of only 'to English' language pairs). Note that $\tau$ we use here is WMT-variant suggested in \cite{wmt-2021-machine}. For fair comparison, in (1), only 'to English' pairs are used. For MBART \cite{liu-etal-2020-multilingual-denoising} we used mbart50-many-to-many model, for PRISM \cite{thompson-post-2020-automatic}, we used m39v1 model.}
\end{table}

\section{Conclusion}
In this study, we propose HaRiM${^+}$ as a new summarization metric, which exploits the power of the summarization model for evaluation accompanied with the hallucination risk into consideration.
For evaluating summaries, HaRiM${^+}$ only requires the summarization model without further training, additional module, or complicated pipelines.
Our method further demonstrates the merit of using summarization models not only for summary generation but also for evaluation.
Throughout the quantitative and qualitative analyses, we show that the HaRiM${^+}$ metric correlates well to human judgment in comprehensive aspects with robust performance, demonstrated with qualitative examples.
We also explored the inductive bias of the model, which emphasizes the importance of training noisy-robust summarization-generation models for evaluation use. 
We leave the potential extension of the metric to another generation task, such as machine translation, as a future remark. 
\bibliographystyle{acl_natbib}
\bibliography{anthology, custom}

\clearpage
\pdfoutput=1
\renewcommand\thefigure{A.\arabic{figure}}    
\setcounter{figure}{0}    
\renewcommand\thetable{B.\arabic{table}}
\setcounter{table}{0}

\appendix

\section{Additional Results}
\subsection{Comparison of HaRiM$^+$ Performance to Reported Values}
We separately represent the meta-evaluation results compared to reported metrics' benchmark scores in Table \ref{tab:reported_comp}. 
Mostly the reported values are using $r$ and $\rho$ to estimate metric performance, which does not fit into our selection of primary means of measure ($\tau$). 
Reason for avoiding the use of $\rho$ is simple: $\rho$ does not guarantee monotonic relation between correlated variables, rather it means linearity, and we found $\tau$ to be more interpretable measure for ranking the quality of article-summary pairs.

\subsection{System-level Metric-Human Correlations on SummEval}
In Table \ref{tab:app_syslv_corr}, we report system-level correlation of metric scores on SummEval benchmark, which contains total 16 systems. To 100 articles, 16 systems (12 abstractive, 4 extractive) present their summary generation.

\subsection{Metric-Human Correlations in Spearman's $r$ and Pearson's $\rho$}
In Table \ref{tab:spearman_pearson}, we provide benchmark results with Spearman's rank coefficient ($r$), and Pearson's $\rho$.
As mentioned earlier, for our set of metric scores, three correlation measures orders almost the same with each other while it is not guaranteed in general.

\subsection{Significance by Randomization Test}
\begin{figure*}
\centering
\includegraphics[width=\textwidth]{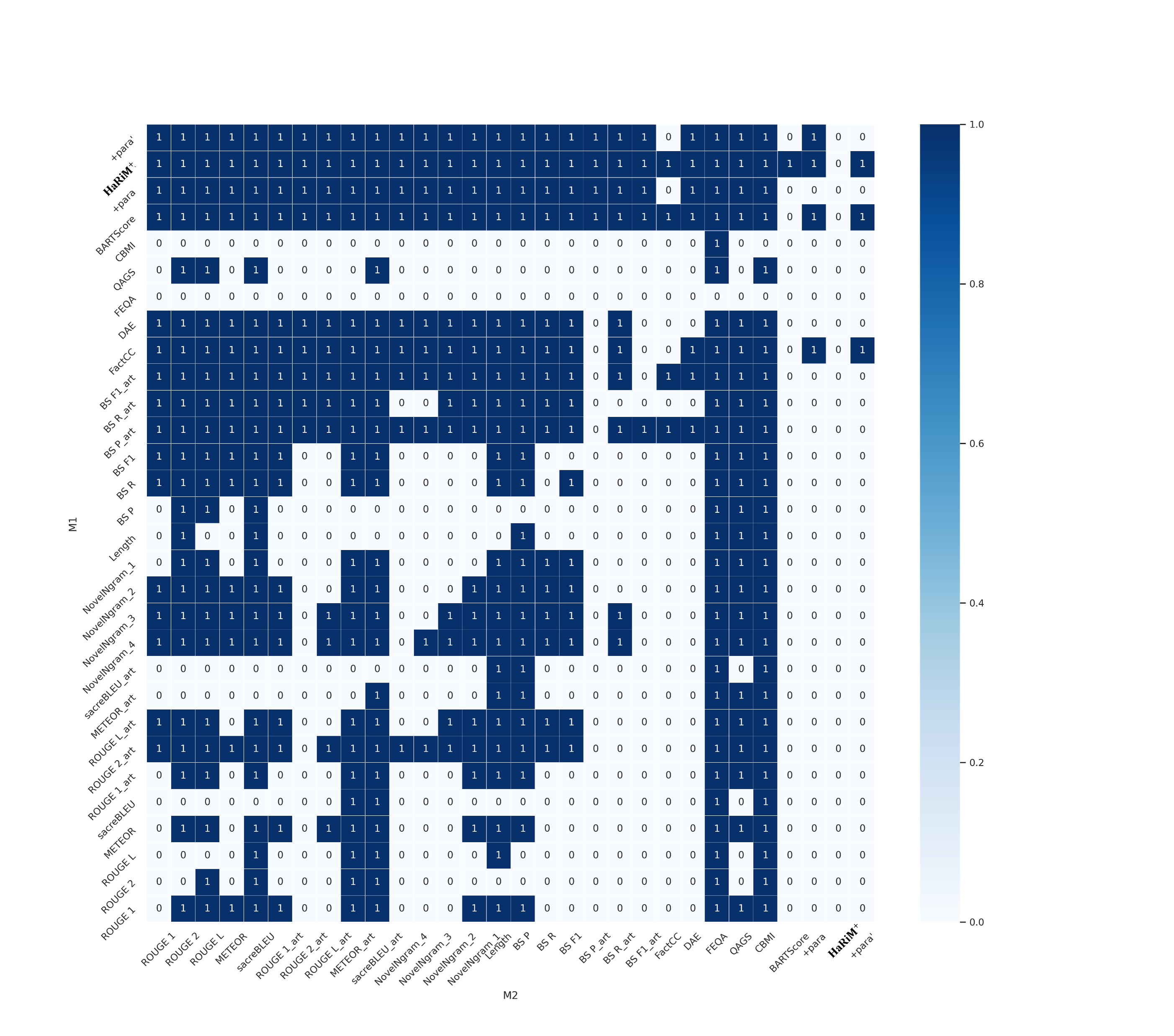}
\caption{\label{fig:FRANK_CNN_permutation} Permutation test done for metric scores on FRANK-CNN/DM. 1 (filled grid) represents significant difference in metric performance, 0 represents negligible difference with confidence >=.95 ($p<=0.05$), i.e. HaRiM is significantly more correlated to human judgements than all the other metrics except itself with a confidence of $>=$95\%. }
\end{figure*}

With randomization test in Figure \ref{fig:FRANK_CNN_permutation}, we can compute the confidence of the difference being coincidant by chance or significant with certain confidence.
We follow the practice of \cite{deutsch-etal-2021-statistical}, PERM-INPUT, as our correlation benchmarking only covers summary-level metric score alignment to human judgement.
We provide randomization test results for every pair of metrics on metric-human correlation on FRANK benchmark, which provides the largest number of metrics are available. 
HaRiM$^+$ largely outperforms the others.

\subsection{Metric-Metric Correlation}
\begin{figure*}
\centering
\includegraphics[width=\textwidth]{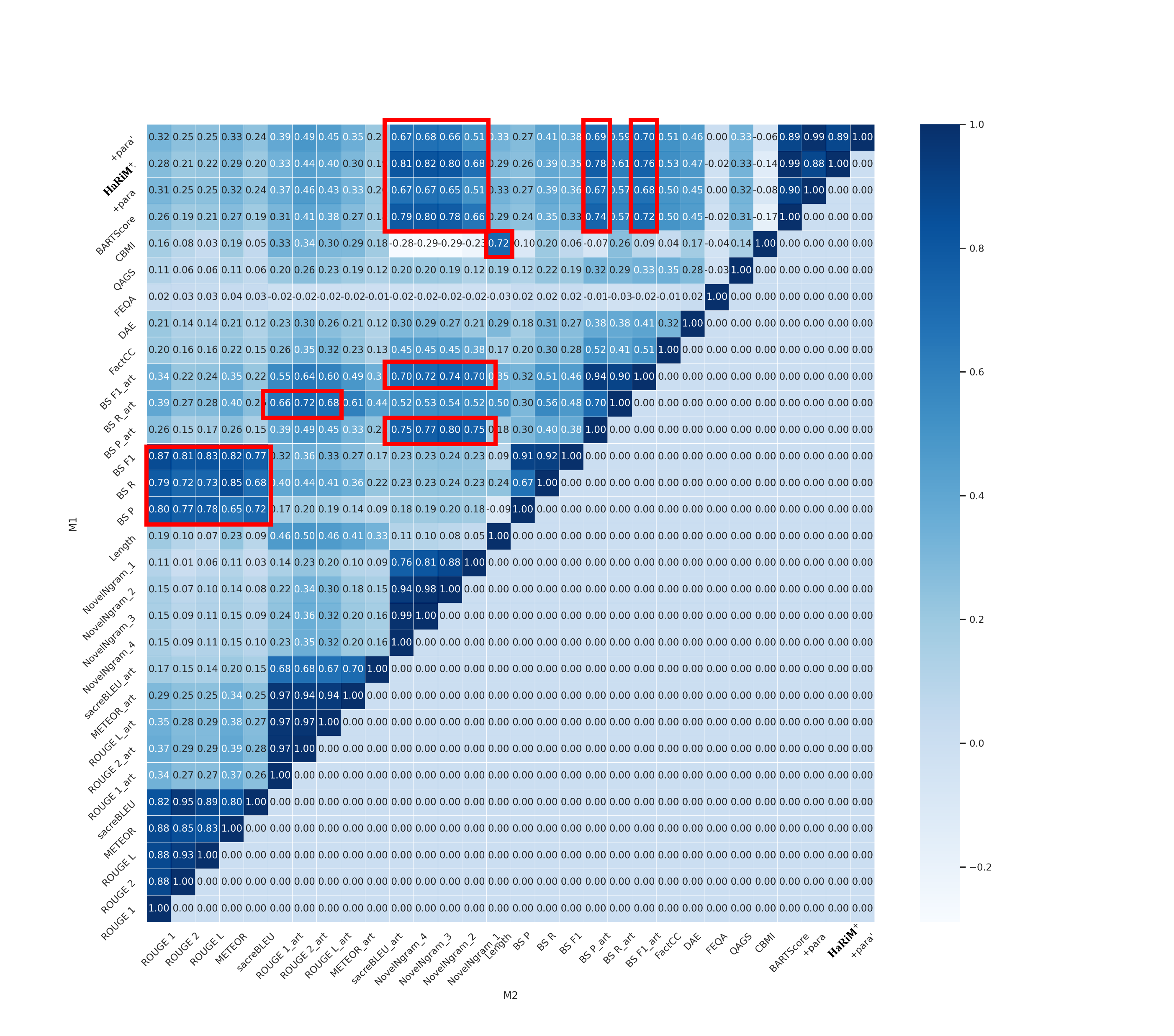}
\caption{\label{fig:metmet_corr_pearson_cnn}Pearson's $\rho$ correlation between metric scores on FRANK-CNN/DM split. The highter the correlation, the similar the metric behavior becomes. Red boxes highlights notable observation which is unexpected behavioral similarity between metrics.}
\end{figure*}

\begin{figure*}
\centering
\includegraphics[width=\textwidth]{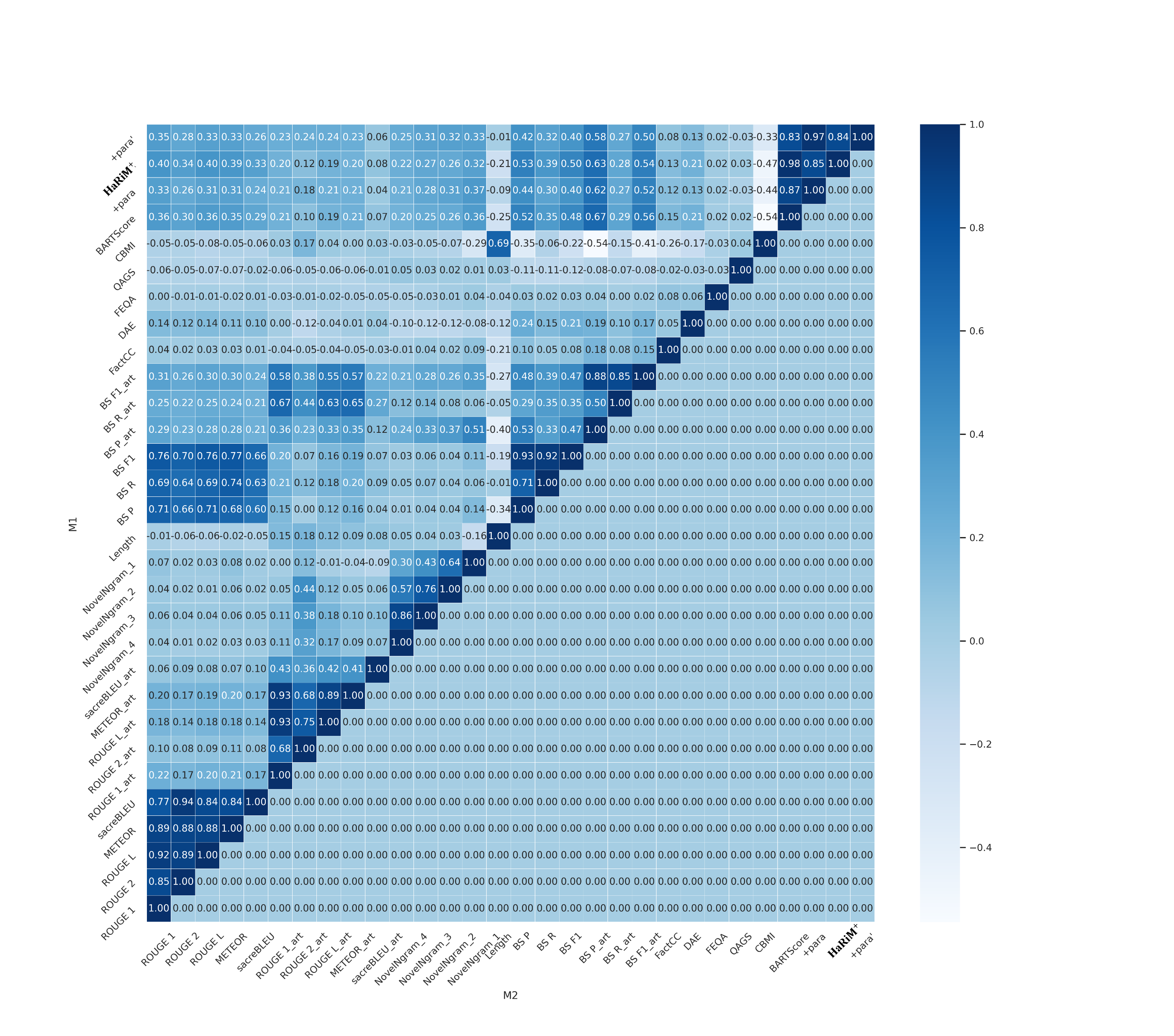}
\caption{\label{fig:metmet_corr_pearson_bbc}Pearson's $\rho$ correlation between metric scores on FRANK-BBC/XSUM split. The highter the correlation, the similar the metric behavior becomes.}
\end{figure*}

In Figure \ref{fig:metmet_corr_pearson_cnn} and \ref{fig:metmet_corr_pearson_bbc}, We provide metric-metric correlation with Pearson\'s $\rho$ which might hint the similarity between metric behaviors.
We highligted several notable trend similarity of the metrics with the red boxes on Figure \ref{fig:metmet_corr_pearson_cnn} according to the following criteria: $\rho$ rounds to .7 or larger, while not a clearly relevant metric (around the diagonal).

Observation shows that text-generation-based metrics correlates well with NovelNgram variants and BERTScore$\_$art (P, F1, not R) while not with ROUGE.
BERTScore behavior differs quite much when applied to article or reference. 
BERTScore measured with reference text resembles behavior of ROUGE scores while they turns more similar to NovelNgrams and text-generation-based metrics (HaRiM$^+$, and BARTScore) for BERTScore-P (BS P$\_$art). 
CBMI, is the most resemblant metric to length of the summary text (L) which records 0.72 in $\rho$.

\subsection{SummEval Separate Results: Abstractive/Extractive System Outputs}
In Figure \ref{tab:absextseparate}, we provide benchmark results ($\tau$ correlation) toward abstractive and extractive summary outputs in separate.
As discussed in the Section \ref{inductivebias}, HaRiM$^+$ correlates better on abstractive system outputs. 

\subsection{More of Qualitative Examples}
We present several more qualitative examples in Table \ref{tab:qual_neg1}, \ref{tab:quala}, \ref{tab:qual_neg2}, \ref{tab:qual_pos1}, and \ref{tab:qual_pos2}.
Those five examples are from FRANK benchmark, three are showcasing hallucinated outputs (Factuality=0) and following two are for factual outputs (Factuality=1).

\section{Analyses}
\subsection{HaRiM variations tested on FRANK}
\label{app:variations}
In Table \ref{tab:vary_marginf}, we show our heuristic trials to aggregate $\Delta=p_{s2s}-p_{lm}$ to make the hallucination risk (HaRiM) better correlate to the human judgements in FRANK benchmark.
We found the original form, denoted as \emph{linear}, works stable than the others.
Applying other function-form (log or exponential) than linear for $\Delta (=p_{s2s}-p{lm})$ was not effective.
Also for aggregating token level scores, we tried applying $\verb|tfidf|$ and $\verb|idf|$, which turned out doing nothing than worsening the correlation as similarly top/bot 5 average do.
Entropy-based scores are also tested but found ineffective.

\subsection{Effect of variables to HaRiM}
We show fine-grained effect of each variables (e.g. $p_{lm},~p_{s2s},~\Delta$) to HaRiM. 
Figure \ref{fig1:replacelm} shows article-summary pair as a datapoint in the plot, here we show each token of the decoded output as a datapoint.
Replacing $p_{lm}$ with empty-sourced decoder inference looks fair even in token-level plot (HaRiM did not change drastically).
HaRiM seems quite dependent on $p_{s2s}$, but as we reported earlier in the main body of this paper (benchmark results), use of $p_{lm}$ quite helps benefits HaRiM$^+$ a lot.

\subsection{Why should not the performance on FRANK benchmark reported with partial correlation}
\label{app:exponly}
The correlation value reported on the Table\ref{tab:qmain}, column FRANK shows correlation to human judgements, not considering partial correlation as suggested in \cite{pagnoni-etal-2021-understanding}. 
A metric, or a scorer for the text-quality measurement does not refer to the system which wrote the text while the partial correlation suggested by \citeauthor{pagnoni-etal-2021-understanding} considers this as a confounding variable that hinders precise meta-evaluation of the metrics.
In Figure \ref{fig:graphicalmodel}, we represent our claim that the generation system should not be taken into account for metric meta-evaluation with two graphical models. 
The graph A shows the view of \citeauthor{pagnoni-etal-2021-understanding}, which considers generation system (i.e. summarization model), into account while the other graph (B) shows ours.
Metric score, $M$, and human judgement, $H$, are both grounded by the $\mathrm{text}$, which blocks the effect of generation system, $S$, in the graphical model; which means considering $S$ for measuring the correlation betweeen $M$ and $H$ is at best doubtful for precise meta-evaluation.

\subsection{SummEval: Why Experts' Annotations not Turkers'?}
 In Figure \ref{fig:spurrious_abs}, and \ref{fig:spurrious_ext}, we plotted averaged experts' annotations over annotators and 4 aspects of quality (i.e. consistency, cohenrence, fluency, relevance), versus turkers' counterpart of those.
 Turkers' judgement of quality in average look irrelevant to correspondings of experts. 
 As mentioned in \cite{fabbri-etal-2021-summeval}, expert annotators are re-instructred after the first round of annotation, which resulted improved inter-annotator-agreement.
 Thus, trusting in annotations from experts but not for crowd-workers of SummEval is plausible as other works done on SummEval benchmark annotation set.

\section{Implementation Details}
\label{app:implementation_details}
\subsection{QAGS}
\textbf{QAGS scorer:} We used original code from the author (\url{https://github.com/W4ngatang/qags}) except its missing part which provide functions for matching the generated answer with GT, in SQuAD style. \\
\noindent\textbf{Aggregating Annotations:} ``Yes" are considered 1 and ``no" considered 0 (coherent to the sign of the FRANK benchmark annotations) to finally obtain averaged factuality label we used. Annotations are also from the original repository.
\subsection{BERTScore}
We used $\verb|BERTScore==0.3.11|$ \\ (\url{https://github.com/Tiiiger/bert_score}) which defaults to RoBERTa-large weight for text.
\subsection{N-gram Metrics}
For traditional N-gram-based metrics, we used huggingface's $\verb|datasets.load_metric()|$ wrapper to load SacreBLEU, METEOR, and ROUGE.
Codebase of each metric is as follow:
\begin{itemize}
  \itemsep0em
    \item SacreBLEU: $\verb|sacreBLEU==2.1.0|$ from the repository (\url{https://github.com/mjpost/sacrebleu}).
    \item METEOR: $\verb|nltk.translate.meteor_score|$ from $\verb|NLTK=3.6.4|$.
    \item ROUGE: We used $\verb|datasets.load_metric('rouge')|$ which uses \url{https://github.com/google-research/google-research/tree/master/rouge} as its codebase.
\end{itemize}
\subsection{Novel Ngram}
Equation \ref{eq:nn} describes our computation of NovelNgram, which does not consider duplication of the tokens.
Minus sign is applied to use it as a higher-is-better score.
\begin{equation}
\label{eq:nn}
\mathrm{NN_{i}} = -\frac{\mathrm{len}(\mathrm{set(Ngram_i^{output})}-\mathrm{set(Ngram_i^{article})})}{\mathrm{len(set(Ngram_i^{article}))}} 
\end{equation}
\subsection{CBMI}
\label{app:cbmi}
Original implementation of conditional bilingual mutual information (CBMI) proposed by \citeauthor{zhang-etal-2022-conditional} uses minibatch statistics for nomalization. 
Instead we take whole examples of FRANK benchmark to compute the CBMI statistics.

\subsection{List of Reused Metric Scores from FRANK repository}
We measured all the other metric scores on all benchmarks other than specified below.
\begin{itemize}
  \itemsep0em
    \item FactCC \cite{kryscinski-etal-2020-evaluating}
    \item Dependency Arc Entailment (Dep Entail) \cite{goyal-durrett-2020-evaluating}
    \item FEQA \cite{durmus-etal-2020-feqa}
    \item QAGS on FRANK benchmark \cite{wang-etal-2020-asking, pagnoni-etal-2021-understanding} \\(on QAGS annotation set, we scored with re-implemented scorer)
\end{itemize}

\begin{table*}[ht]
\resizebox{\textwidth}{!}{%
\begin{tabular}{l|cc|cc|cc}
                             & \multicolumn{2}{l|}{QAGS-CNNDM} & \multicolumn{2}{l|}{QAGS-XSUM}  & \multicolumn{2}{l}{SummEval (1200 outputs)} \\ \cline{2-7} 
                             & $r$            & $\rho$         & $r$            & $\rho$         & $r$                  & $\rho$               \\ \hline
QAGS                         & 0.382          & 0.466          & 0.203          & {\ul 0.217}    &                      &                      \\
FFCI$\_${BERTScore}*         & 0.485          & 0.486          & {\ul 0.200}    & 0.190          & 0.285                & 0.308                \\
QuestEval$\_${F1}*           & 0.492          & 0.445          & 0.007          & 0.010          & 0.370                & 0.339                \\ \hline
CoCo$\_$span*                  & 0.573          & 0.501          & 0.187          & 0.187          & \textbf{0.436}       & 0.410                \\
CoCo$\_$sent*                  & {\ul 0.588}    & 0.523          & \textbf{0.241} & \textbf{0.227} & 0.420                & 0.390                \\ \hline
HaRiM$^+$ (BART-large+cnn+para) & 0.530          & {\ul 0.610}    &                &                & 0.405                & 0.430                \\
HaRiM$^+$ (BART-large+cnn)      & \textbf{0.620} & \textbf{0.679} &                &                & 0.392                & 0.415                \\
HaRiM$^+$ (BRIO)                & 0.514          & 0.569          &                &                & 0.417                & \textbf{0.443}      
\end{tabular}%
}
\caption{\label{tab:reported_comp}Metric correlation to human judgements on SummEval-abstractive (1200 out of 1600 total examples) QAGS annotation set in Pearson's $\rho$ and Spearman $r$. * notes that the values are copied from each paper \cite{xie-etal-2021-factual-consistency}.}
\end{table*}

\begin{table*}[]
\resizebox{\textwidth}{!}{%
\begin{tabular}{lllllllllllll}
\hline
\multicolumn{1}{l|}{}                                        & \multicolumn{12}{c}{\textbf{SummEval   (system-level correlation, 16 systems)}}                                                                                                                                                                                                                                                                                                                                                          \\ \hline
\multicolumn{1}{l|}{{\color[HTML]{333333} \textbf{}}}        & \multicolumn{3}{c|}{\textbf{consistency}}                                                                 & \multicolumn{3}{c|}{\textbf{coherence}}                                                                   & \multicolumn{3}{c|}{\textbf{fluency}}                                                                     & \multicolumn{3}{c}{\textbf{relevance}}                                                               \\
\multicolumn{1}{l|}{\textbf{Metrics}}                        & \multicolumn{1}{c}{\textbf{$\tau$}} & \multicolumn{1}{c}{\textbf{$\rho$}} & \multicolumn{1}{c|}{\textbf{$r$}}     & \multicolumn{1}{c}{\textbf{$\tau$}} & \multicolumn{1}{c}{\textbf{$\rho$}} & \multicolumn{1}{c|}{\textbf{$r$}}     & \multicolumn{1}{c}{\textbf{$\tau$}} & \multicolumn{1}{c}{\textbf{$\rho$}} & \multicolumn{1}{c|}{\textbf{$r$}}     & \multicolumn{1}{c}{\textbf{$\tau$}} & \multicolumn{1}{c}{\textbf{$\rho$}} & \multicolumn{1}{c}{\textbf{$r$}} \\ \hline
\multicolumn{13}{l}{\textbf{n-gram-matching}}                                                                                                                                                                                                                                                                                                                                                                                                                                                           \\ \hline
\multicolumn{1}{l|}{ROUGE 1}                                 & 0.500                            & 0.662                            & \multicolumn{1}{l|}{0.688}          & 0.267                            & 0.063                            & \multicolumn{1}{l|}{0.459}          & 0.450                            & 0.554                            & \multicolumn{1}{l|}{0.635}          & 0.500                            & 0.550                            & 0.682                          \\
\multicolumn{1}{l|}{ROUGE 2}                                 & 0.600                            & 0.653                            & \multicolumn{1}{l|}{0.765}          & 0.233                            & 0.085                            & \multicolumn{1}{l|}{0.338}          & 0.483                            & 0.542                            & \multicolumn{1}{l|}{0.676}          & 0.433                            & 0.561                            & 0.626                          \\
\multicolumn{1}{l|}{ROUGE L}                                 & 0.283                            & {\ul 0.697}                      & \multicolumn{1}{l|}{0.385}          & 0.383                            & 0.204                            & \multicolumn{1}{l|}{0.506}          & 0.467                            & 0.624                            & \multicolumn{1}{l|}{0.600}          & 0.517                            & 0.600                            & 0.712                          \\
\multicolumn{1}{l|}{METEOR}                                  & 0.550                            & 0.559                            & \multicolumn{1}{l|}{0.703}          & 0.017                            & 0.044                            & \multicolumn{1}{l|}{0.026}          & 0.267                            & 0.449                            & \multicolumn{1}{l|}{0.385}          & 0.250                            & 0.438                            & 0.312                          \\
\multicolumn{1}{l|}{sacreBLEU}                               & -0.050                           & 0.175                            & \multicolumn{1}{l|}{-0.118}         & 0.383                            & 0.493                            & \multicolumn{1}{l|}{0.529}          & 0.233                            & 0.233                            & \multicolumn{1}{l|}{0.318}          & 0.283                            & 0.462                            & 0.418                          \\
\multicolumn{1}{l|}{ROUGE 1$\_$art}                            & 0.467                            & 0.467                            & \multicolumn{1}{l|}{0.626}          & 0.000                            & 0.028                            & \multicolumn{1}{l|}{-0.068}         & 0.217                            & 0.375                            & \multicolumn{1}{l|}{0.288}          & 0.200                            & 0.324                            & 0.174                          \\
\multicolumn{1}{l|}{ROUGE 2$\_$art}                            & 0.500                            & 0.599                            & \multicolumn{1}{l|}{0.688}          & 0.067                            & 0.072                            & \multicolumn{1}{l|}{-0.026}         & 0.283                            & 0.515                            & \multicolumn{1}{l|}{0.329}          & 0.267                            & 0.370                            & 0.212                          \\
\multicolumn{1}{l|}{ROUGE L$\_$art}                            & 0.550                            & 0.618                            & \multicolumn{1}{l|}{0.726}          & 0.117                            & 0.164                            & \multicolumn{1}{l|}{0.018}          & 0.300                            & 0.541                            & \multicolumn{1}{l|}{0.362}          & 0.317                            & 0.421                            & 0.265                          \\
\multicolumn{1}{l|}{METEOR$\_$art}                             & 0.467                            & 0.513                            & \multicolumn{1}{l|}{0.621}          & 0.000                            & 0.082                            & \multicolumn{1}{l|}{-0.021}         & 0.250                            & 0.430                            & \multicolumn{1}{l|}{0.335}          & 0.233                            & 0.397                            & 0.226                          \\
\multicolumn{1}{l|}{sacreBLEU$\_$art}                          & 0.450                            & 0.287                            & \multicolumn{1}{l|}{0.621}          & 0.083                            & 0.299                            & \multicolumn{1}{l|}{0.176}          & 0.200                            & 0.277                            & \multicolumn{1}{l|}{0.318}          & 0.183                            & 0.351                            & 0.209                          \\ \hline
\multicolumn{13}{l}{\textbf{N-gram stats}}                                                                                                                                                                                                                                                                                                                                                                                                                                                              \\ \hline
\multicolumn{1}{l|}{NovelNgram$\_$4}                           & 0.400                            & 0.623                            & \multicolumn{1}{l|}{0.553}          & 0.300                            & 0.704                            & \multicolumn{1}{l|}{0.435}          & 0.450                            & {\ul 0.691}                      & \multicolumn{1}{l|}{0.606}          & 0.367                            & 0.664                            & 0.506                          \\
\multicolumn{1}{l|}{NovelNgram$\_$3}                           & 0.367                            & 0.590                            & \multicolumn{1}{l|}{0.512}          & 0.333                            & 0.657                            & \multicolumn{1}{l|}{0.453}          & 0.417                            & 0.649                            & \multicolumn{1}{l|}{0.594}          & 0.367                            & 0.631                            & 0.506                          \\
\multicolumn{1}{l|}{NovelNgram$\_$2}                           & 0.300                            & 0.464                            & \multicolumn{1}{l|}{0.444}          & 0.367                            & 0.615                            & \multicolumn{1}{l|}{0.524}          & 0.417                            & 0.522                            & \multicolumn{1}{l|}{0.576}          & 0.400                            & 0.570                            & 0.541                          \\
\multicolumn{1}{l|}{NovelNgram$\_$1}                           & -0.017                           & 0.016                            & \multicolumn{1}{l|}{0.006}          & 0.417                            & 0.456                            & \multicolumn{1}{l|}{0.529}          & 0.167                            & 0.091                            & \multicolumn{1}{l|}{0.241}          & 0.183                            & 0.276                            & 0.244                          \\
\multicolumn{1}{l|}{Length (no. tokens)}                     & 0.417                            & 0.348                            & \multicolumn{1}{l|}{0.571}          & -0.050                           & -0.009                           & \multicolumn{1}{l|}{-0.112}         & 0.200                            & 0.262                            & \multicolumn{1}{l|}{0.268}          & 0.183                            & 0.239                            & 0.156                          \\ \hline
\multicolumn{13}{l}{\textbf{Contextual Embedding}}                                                                                                                                                                                                                                                                                                                                                                                                                                                      \\ \hline
\multicolumn{1}{l|}{BERTScore P}                             & -0.233                           & -0.254                           & \multicolumn{1}{l|}{-0.341}         & 0.300                            & 0.457                            & \multicolumn{1}{l|}{0.406}          & 0.017                            & -0.122                           & \multicolumn{1}{l|}{0.047}          & 0.067                            & 0.126                            & 0.150                          \\
\multicolumn{1}{l|}{BERTScore R}                             & 0.617                            & 0.459                            & \multicolumn{1}{l|}{0.809}          & 0.550                            & 0.671                            & \multicolumn{1}{l|}{0.697}          & 0.600                            & 0.486                            & \multicolumn{1}{l|}{{\ul 0.806}}    & 0.617                            & 0.749                            & {\ul 0.797}                    \\
\multicolumn{1}{l|}{BERTScore F1}                            & 0.017                            & -0.039                           & \multicolumn{1}{l|}{0.021}          & 0.550                            & 0.623                            & \multicolumn{1}{l|}{0.715}          & 0.333                            & 0.083                            & \multicolumn{1}{l|}{0.432}          & 0.417                            & 0.373                            & 0.497                          \\
\multicolumn{1}{l|}{BERTScore P$\_$art}                        & \textbf{0.750}                   & 0.623                            & \multicolumn{1}{l|}{\textbf{0.903}} & 0.317                            & 0.441                            & \multicolumn{1}{l|}{0.453}          & 0.567                            & 0.589                            & \multicolumn{1}{l|}{0.756}          & 0.517                            & 0.653                            & 0.676                          \\
\multicolumn{1}{l|}{BERTScore R$\_$art}                        & 0.583                            & 0.654                            & \multicolumn{1}{l|}{0.809}          & 0.450                            & 0.715                            & \multicolumn{1}{l|}{0.559}          & 0.500                            & 0.691                            & \multicolumn{1}{l|}{0.662}          & 0.550                            & 0.714                            & 0.635                          \\
\multicolumn{1}{l|}{BERTScore F1$\_$art}                       & {\ul 0.683}                      & 0.680                            & \multicolumn{1}{l|}{{\ul 0.868}}    & 0.417                            & 0.623                            & \multicolumn{1}{l|}{0.559}          & {\ul 0.600}                      & 0.684                            & \multicolumn{1}{l|}{0.753}          & 0.583                            & 0.727                            & 0.691                          \\ \hline
\multicolumn{13}{l}{\textbf{Text Generation based}}                                                                                                                                                                                                                                                                                                                                                                                                                                                     \\ \hline
\multicolumn{1}{l|}{CBMI (BART$\_$base + cnn)*}                & 0.433                            & 0.483                            & \multicolumn{1}{l|}{0.632}          & -0.033                           & -0.119                           & \multicolumn{1}{l|}{-0.132}         & 0.217                            & 0.384                            & \multicolumn{1}{l|}{0.238}          & 0.200                            & 0.185                            & 0.132                          \\
\multicolumn{1}{l|}{BARTScore (BART-large + cnn)**}          & 0.183                            & 0.301                            & \multicolumn{1}{l|}{0.259}          & {\ul 0.717}                      & 0.812                            & \multicolumn{1}{l|}{{\ul 0.871}}    & 0.467                            & 0.423                            & \multicolumn{1}{l|}{0.559}          & 0.550                            & 0.592                            & 0.621                          \\
\multicolumn{1}{l|}{BARTScore (BART-large + cnn +   para)**} & 0.283                            & 0.577                            & \multicolumn{1}{l|}{0.424}          & 0.650                            & \textbf{0.891}                   & \multicolumn{1}{l|}{0.809}          & 0.567                            & 0.687                            & \multicolumn{1}{l|}{0.735}          & 0.617                            & {\ul 0.783}                      & 0.750                          \\ \hline
\multicolumn{13}{l}{\textbf{Proposed}}                                                                                                                                                                                                                                                                                                                                                                                                                                                                  \\ \hline
\multicolumn{1}{l|}{\textbf{HaRiM+ (BART$\_$large + cnn)}}     & 0.250                            & 0.492                            & \multicolumn{1}{l|}{0.368}          & \textbf{0.817}                   & 0.835                            & \multicolumn{1}{l|}{\textbf{0.926}} & 0.500                            & 0.593                            & \multicolumn{1}{l|}{0.679}          & {\ul 0.650}                      & 0.721                            & 0.756                          \\
\multicolumn{1}{l|}{HaRiM+ (BART$\_$large + cnn +   para)}     & 0.383                            & \textbf{0.701}                   & \multicolumn{1}{l|}{0.562}          & 0.617                            & {\ul 0.860}                      & \multicolumn{1}{l|}{0.762}          & \textbf{0.667}                   & \textbf{0.790}                   & \multicolumn{1}{l|}{\textbf{0.859}} & \textbf{0.717}                   & \textbf{0.851}                   & \textbf{0.859} \\ \hline                
\end{tabular}%
}
\caption{System-level correlation on SummEval, total 16 systems (12 abstractive, 4 extractive). \textbf{Bold}face numbers represent the best and underlined are the second-best. We omit abstractive-systems-only result as its trend is similar to above.}
\label{tab:app_syslv_corr}
\end{table*}

\begin{table*}[]
\centering
\resizebox{\textwidth}{!}{%
\begin{tabular}{lcccccccccccccccc}
\hline
\multicolumn{1}{l|}{}  & \multicolumn{4}{c|}{\textbf{CNNDM}}   & \multicolumn{1}{l}{} & \multicolumn{1}{l}{} & \multicolumn{1}{l}{} & \multicolumn{1}{l}{} & \multicolumn{1}{l}{} & \multicolumn{1}{l}{} & \multicolumn{1}{l}{} & \multicolumn{1}{l|}{}  & \multicolumn{4}{c}{\textbf{XSUM}}   \\ \hline
\multicolumn{1}{l|}{}  & \multicolumn{2}{c}{\textbf{FRANK}} & \multicolumn{2}{c|}{\textbf{QAGS}}  & \multicolumn{8}{c|}{\textbf{SummEval}}       & \multicolumn{2}{c}{\textbf{FRANK}} & \multicolumn{2}{c}{\textbf{QAGS}}  \\ \hline
\multicolumn{1}{l|}{}  & \multicolumn{2}{c}{\textbf{Factuality}}  & \multicolumn{2}{c|}{\textbf{Factuality}}  & \multicolumn{2}{c}{\textbf{con}} & \multicolumn{2}{c}{\textbf{coh}} & \multicolumn{2}{c}{\textbf{flu}} & \multicolumn{2}{c|}{\textbf{rel}}   & \multicolumn{2}{c}{\textbf{Factuality}}  & \multicolumn{2}{c}{\textbf{Factuality}}  \\ \hline
\multicolumn{1}{l|}{\textbf{Metrics}} & \textbf{$r$}  & \multicolumn{1}{c|}{\textbf{$\rho$}}  & \textbf{$r$}  & \multicolumn{1}{c|}{\textbf{$\rho$}}  & \textbf{$r$} & \multicolumn{1}{c|}{\textbf{$\rho$}} & \textbf{$r$} & \multicolumn{1}{c|}{\textbf{$\rho$}} & \textbf{$r$} & \multicolumn{1}{c|}{\textbf{$\rho$}} & \textbf{$r$} & \multicolumn{1}{c|}{\textbf{$\rho$}}  & \textbf{$r$}  & \multicolumn{1}{c|}{\textbf{$\rho$}}  & \textbf{$r$}  & \multicolumn{1}{c|}{\textbf{$\rho$}}  \\ \hline
\textbf{n-gram-matching}   & \multicolumn{1}{l}{} & \multicolumn{1}{l}{} & \multicolumn{1}{l}{} & \multicolumn{1}{l}{} & \multicolumn{1}{l}{\textbf{}} & \multicolumn{1}{l}{\textbf{}} & \multicolumn{1}{l}{\textbf{}} & \multicolumn{1}{l}{\textbf{}} & \multicolumn{1}{l}{\textbf{}} & \multicolumn{1}{l}{\textbf{}} & \multicolumn{1}{l}{\textbf{}} & \multicolumn{1}{l}{\textbf{}}  & \multicolumn{1}{l}{} & \multicolumn{1}{l}{} & \multicolumn{1}{l}{} & \multicolumn{1}{l}{} \\ \hline
\multicolumn{1}{l|}{ROUGE 1}  & 0.239 & \multicolumn{1}{c|}{0.254} & -0.072  & \multicolumn{1}{c|}{-0.013}  & 0.167  & \multicolumn{1}{c|}{0.133}  & 0.181  & \multicolumn{1}{c|}{0.175}  & 0.136  & \multicolumn{1}{c|}{0.080}  & 0.323  & \multicolumn{1}{c|}{0.289} & 0.153 & \multicolumn{1}{c|}{0.179} & 0.148 & 0.163 \\
\multicolumn{1}{l|}{ROUGE 2}  & 0.178 & \multicolumn{1}{c|}{0.181} & -0.151  & \multicolumn{1}{c|}{-0.019}  & 0.147  & \multicolumn{1}{c|}{0.128}  & 0.131  & \multicolumn{1}{c|}{0.138}  & 0.087  & \multicolumn{1}{c|}{0.062}  & 0.240  & \multicolumn{1}{c|}{0.234} & 0.154 & \multicolumn{1}{c|}{0.186} & 0.134 & 0.145 \\
\multicolumn{1}{l|}{ROUGE L}  & 0.186 & \multicolumn{1}{c|}{0.194} & -0.100  & \multicolumn{1}{c|}{-0.042}  & 0.142  & \multicolumn{1}{c|}{0.115}  & 0.155  & \multicolumn{1}{c|}{0.160}  & 0.110  & \multicolumn{1}{c|}{0.079}  & 0.248  & \multicolumn{1}{c|}{0.231} & 0.144 & \multicolumn{1}{c|}{0.182} & 0.121 & 0.117 \\
\multicolumn{1}{l|}{METEOR} & 0.260 & \multicolumn{1}{c|}{0.268} & 0.074 & \multicolumn{1}{c|}{0.050} & 0.173  & \multicolumn{1}{c|}{0.158}  & 0.168  & \multicolumn{1}{c|}{0.165}  & 0.114  & \multicolumn{1}{c|}{0.091}  & 0.360  & \multicolumn{1}{c|}{0.312} & 0.148 & \multicolumn{1}{c|}{0.165} & 0.156 & 0.157 \\
\multicolumn{1}{l|}{sacreBLEU}  & 0.179 & \multicolumn{1}{c|}{0.169} & -0.116  & \multicolumn{1}{c|}{-0.063}  & 0.117  & \multicolumn{1}{c|}{0.102}  & 0.250  & \multicolumn{1}{c|}{0.238}  & 0.139  & \multicolumn{1}{c|}{0.113}  & 0.290  & \multicolumn{1}{c|}{0.290} & 0.139 & \multicolumn{1}{c|}{0.156} & 0.016 & 0.036 \\
\multicolumn{1}{l|}{ROUGE 1$\_$art} & 0.244 & \multicolumn{1}{c|}{0.255} & 0.336 & \multicolumn{1}{c|}{0.355} & 0.137  & \multicolumn{1}{c|}{0.142}  & 0.074  & \multicolumn{1}{c|}{0.049}  & 0.087  & \multicolumn{1}{c|}{0.075}  & 0.209  & \multicolumn{1}{c|}{0.179} & -0.004  & \multicolumn{1}{c|}{-0.017}  & -0.103  & -0.065  \\
\multicolumn{1}{l|}{ROUGE 2$\_$art} & 0.327 & \multicolumn{1}{c|}{0.331} & 0.427 & \multicolumn{1}{c|}{0.475} & 0.252  & \multicolumn{1}{c|}{0.247}  & 0.123  & \multicolumn{1}{c|}{0.099}  & 0.188  & \multicolumn{1}{c|}{0.154}  & 0.245  & \multicolumn{1}{c|}{0.215} & 0.033 & \multicolumn{1}{c|}{0.012} & 0.091 & 0.107 \\
\multicolumn{1}{l|}{ROUGE L$\_$art} & 0.296 & \multicolumn{1}{c|}{0.297} & 0.411 & \multicolumn{1}{c|}{0.462} & 0.242  & \multicolumn{1}{c|}{0.258}  & 0.155  & \multicolumn{1}{c|}{0.133}  & 0.177  & \multicolumn{1}{c|}{0.159}  & 0.252  & \multicolumn{1}{c|}{0.230} & 0.012 & \multicolumn{1}{c|}{0.000} & -0.024  & 0.014 \\
\multicolumn{1}{l|}{METEOR$\_$art}  & 0.229 & \multicolumn{1}{c|}{0.230} & 0.324 & \multicolumn{1}{c|}{0.277} & 0.122  &\multicolumn{1}{c|}{ 0.143}  & 0.053  & \multicolumn{1}{c|}{0.011}  & 0.093  & \multicolumn{1}{c|}{0.091}  & 0.150  & \multicolumn{1}{c|}{0.129} & 0.005 & \multicolumn{1}{c|}{-0.005}  & -0.071  & -0.015  \\
\multicolumn{1}{l|}{sacreBLEU$\_$art} & 0.202 & \multicolumn{1}{c|}{0.093} & 0.337 & \multicolumn{1}{c|}{0.180} & 0.073  & \multicolumn{1}{c|}{0.117}  & 0.124  & \multicolumn{1}{c|}{0.059}  & 0.071  & \multicolumn{1}{c|}{0.045}  & 0.127  & \multicolumn{1}{c|}{0.184} & -0.046  & \multicolumn{1}{c|}{-0.042}  & -0.186  & 0.047 \\ \hline
\textbf{N-gram stats}  & & & &   & \textbf{}  & \textbf{}  & \textbf{}  & \textbf{}  &  &  & \textbf{}  & \textbf{}  & & & & \\ \hline
\multicolumn{1}{l|}{NovelNgram$\_$4}  & 0.358 & \multicolumn{1}{c|}{0.386} & 0.516 & \multicolumn{1}{c|}{0.600} & 0.277  & \multicolumn{1}{c|}{0.280}  & 0.295  & \multicolumn{1}{c|}{0.283}  & -0.231 & \multicolumn{1}{c|}{-0.221} & 0.282  & \multicolumn{1}{c|}{0.285} & 0.018 & \multicolumn{1}{c|}{0.088} & 0.073 & 0.107 \\
\multicolumn{1}{l|}{NovelNgram$\_$3}  & 0.355 & \multicolumn{1}{c|}{0.390} & 0.494 & \multicolumn{1}{c|}{0.591} & 0.290  & \multicolumn{1}{c|}{0.276}  & 0.300  & \multicolumn{1}{c|}{0.291}  & -0.235 & \multicolumn{1}{c|}{-0.219} & 0.286  & \multicolumn{1}{c|}{0.289} & 0.071 & \multicolumn{1}{c|}{0.105} & 0.107 & 0.118 \\
\multicolumn{1}{l|}{NovelNgram$\_$2}  & 0.337 & \multicolumn{1}{c|}{0.384} & 0.439 & \multicolumn{1}{c|}{0.570} & 0.276  & \multicolumn{1}{c|}{0.252}  & 0.298  & \multicolumn{1}{c|}{0.292}  & -0.208 & \multicolumn{1}{c|}{-0.191} & 0.283  & \multicolumn{1}{c|}{0.287} & 0.064 & \multicolumn{1}{c|}{0.093} & {\ul 0.170} & 0.156 \\
\multicolumn{1}{l|}{NovelNgram$\_$1}  & 0.286 & \multicolumn{1}{c|}{0.349} & 0.282 & \multicolumn{1}{c|}{0.410} & 0.123  & \multicolumn{1}{c|}{0.114}  & 0.271  & \multicolumn{1}{c|}{0.267}  & -0.070 & \multicolumn{1}{c|}{-0.087} & 0.229  & \multicolumn{1}{c|}{0.242} & 0.111 & \multicolumn{1}{c|}{0.119} & 0.158 & {\ul 0.178} \\
\multicolumn{1}{l|}{Length (no. tokens)}  & 0.247 & \multicolumn{1}{c|}{0.207} & 0.263 & \multicolumn{1}{c|}{0.277} & 0.096  & \multicolumn{1}{c|}{0.099}  & 0.048  & \multicolumn{1}{c|}{0.044}  & -0.008 & \multicolumn{1}{c|}{0.004}  & 0.230  & \multicolumn{1}{c|}{0.208} & -0.133  & \multicolumn{1}{c|}{-0.144}  & -0.171  & -0.184  \\ \hline
\textbf{Contextual Embedding} & & & &   &  &  &  &  &  &  &  &   & & & & \\ \hline
\multicolumn{1}{l|}{BERTScore P}  & 0.221 & \multicolumn{1}{c|}{0.237} & -0.095  & \multicolumn{1}{c|}{-0.051}  & 0.049  & \multicolumn{1}{c|}{0.052}  & 0.336  & \multicolumn{1}{c|}{0.320}  & 0.152  & \multicolumn{1}{c|}{0.125}  & 0.245  & \multicolumn{1}{c|}{0.266} & \textbf{0.186}  & \multicolumn{1}{c|}{\textbf{0.208}}  & 0.022 & 0.030 \\
\multicolumn{1}{l|}{BERTScore R}  & 0.327 & \multicolumn{1}{c|}{0.360} & 0.026 & \multicolumn{1}{c|}{0.015} & 0.171  & \multicolumn{1}{c|}{0.158}  & 0.335  & \multicolumn{1}{c|}{0.340}  & 0.139  & \multicolumn{1}{c|}{0.126}  & 0.426  & \multicolumn{1}{c|}{0.415} & 0.131 & \multicolumn{1}{c|}{0.135} & 0.078 & 0.095 \\
\multicolumn{1}{l|}{BERTScore F1} & 0.304 & \multicolumn{1}{c|}{0.329} & -0.041  & \multicolumn{1}{c|}{-0.020}  & 0.107  & \multicolumn{1}{c|}{0.100}  & 0.378  & \multicolumn{1}{c|}{0.375}  & 0.167  & \multicolumn{1}{c|}{0.144}  & 0.360  & \multicolumn{1}{c|}{0.367} & 0.174 & \multicolumn{1}{c|}{0.186} & 0.049 & 0.072 \\
\multicolumn{1}{l|}{BERTScore P$\_$art} & 0.465 & \multicolumn{1}{c|}{0.513} & 0.493 & \multicolumn{1}{c|}{0.548} & {\ul 0.350}  & \multicolumn{1}{c|}{{\ul 0.338}}  & 0.449  & \multicolumn{1}{c|}{{\ul 0.429}}  & {\ul 0.351}  & \multicolumn{1}{c|}{0.300}  & {\ul 0.443}  & \multicolumn{1}{c|}{{\ul 0.422}} & 0.176 & \multicolumn{1}{c|}{{\ul 0.196}} & -0.028  & -0.026  \\
\multicolumn{1}{l|}{BERTScore R$\_$art} & 0.395 & \multicolumn{1}{c|}{0.426} & 0.452 & \multicolumn{1}{c|}{0.497} & 0.175  & \multicolumn{1}{c|}{0.180}  & 0.230  & \multicolumn{1}{c|}{0.215}  & 0.180  &\multicolumn{1}{c|}{ 0.145}  & 0.344  & \multicolumn{1}{c|}{0.326} & 0.046 & \multicolumn{1}{c|}{0.069} & -0.049  & -0.053  \\
\multicolumn{1}{l|}{BERTScore F1$\_$art}  & 0.464 & \multicolumn{1}{c|}{0.514} & 0.493 & \multicolumn{1}{c|}{0.556} & 0.295  & \multicolumn{1}{c|}{0.292}  & 0.381  & \multicolumn{1}{c|}{0.358}  & 0.299  & \multicolumn{1}{c|}{0.246}  & \textbf{0.447} & \multicolumn{1}{c|}{\textbf{0.423}} & 0.137 &\multicolumn{1}{c|}{ 0.157} & -0.054  & -0.048  \\ \hline
\textbf{Neural entailment}  & & & &   &  &  &  &  &  &  &  &   & & & & \\ \hline
\multicolumn{1}{l|}{FactCC} & 0.438 & \multicolumn{1}{c|}{0.492} & & \multicolumn{1}{c|}{}  &  &  &  &  &  &  &  & \multicolumn{1}{c|}{}  & 0.072 & \multicolumn{1}{c|}{0.072} & & \\
\multicolumn{1}{l|}{Dep Entail} & 0.447 & \multicolumn{1}{c|}{0.440} & & \multicolumn{1}{c|}{}  &  &  &  &  &  &  &  & \multicolumn{1}{c|}{}  & 0.113 & \multicolumn{1}{c|}{0.058} & & \\ \hline
\textbf{Q\&A based}  & & & &   &  &  &  &  &  &  &  &   & & & & \\ \hline
\multicolumn{1}{l|}{FEQA}  & -0.010  & \multicolumn{1}{c|}{-0.018}  & & \multicolumn{1}{c|}{}  &  &  &  &  &  &  &  & \multicolumn{1}{c|}{}  & 0.008 & \multicolumn{1}{c|}{0.026} & & \\
\multicolumn{1}{l|}{QAGS}  & 0.267 & \multicolumn{1}{c|}{0.314} & 0.382 & \multicolumn{1}{c|}{0.466} &  &  &  &  &  &  &  & \multicolumn{1}{c|}{}  & -0.007  & \multicolumn{1}{c|}{-0.022}  & \textbf{0.203}  & \textbf{0.217}  \\
\multicolumn{1}{l|}{QAEval-F1 \cite{deutsch-etal-2021-towards}}  &  & \multicolumn{1}{c|}{} &  & \multicolumn{1}{c|}{} &  &  &  &  &  &  & .300 & \multicolumn{1}{c|}{.290}  &   & \multicolumn{1}{c|}{}  &   &   \\\hline
\textbf{Text Generation based}  & & & &   &  &  &  &  &  &  &  &   & & & & \\ \hline
\multicolumn{1}{l|}{CBMI (BART$\_$base + cnn)*} & 0.076 & \multicolumn{1}{c|}{0.099} & 0.040 & \multicolumn{1}{c|}{0.133} & 0.222  & \multicolumn{1}{c|}{0.194}  & -0.013 & \multicolumn{1}{c|}{-0.045} & 0.082  & \multicolumn{1}{c|}{0.030}  & 0.103  & \multicolumn{1}{c|}{0.069} & -0.095  & \multicolumn{1}{c|}{-0.113}  & -0.058  & -0.022  \\
	\multicolumn{1}{l|}{BARTScore (BART-large + cnn)**} & {\ul 0.530} & \multicolumn{1}{c|}{{\ul 0.561}} & {\ul 0.613} & \multicolumn{1}{c|}{{\ul 0.673}} &  0.262  & \multicolumn{1}{c|}{0.249}  & {\ul 0.459}  & \multicolumn{1}{c|}{{\ul 0.429}}  & 0.278  & \multicolumn{1}{c|}{0.231}  & 0.390  & \multicolumn{1}{c|}{0.363} & 0.168 & \multicolumn{1}{c|}{0.174} & 0.097 & 0.080 \\
\multicolumn{1}{l|}{BARTScore (BART-large + cnn + para)**} & 0.507 & \multicolumn{1}{c|}{0.543} & 0.548 & \multicolumn{1}{c|}{0.624} & 0.343  & \multicolumn{1}{c|}{0.328}  & 0.438  & \multicolumn{1}{c|}{0.419}  & 0.350  & \multicolumn{1}{c|}{{\ul 0.305}}  & 0.422  & \multicolumn{1}{c|}{ 0.385} & {\ul 0.177} & \multicolumn{1}{c|}{0.175} & 0.041 & 0.046 \\ \hline
\textbf{Proposed}  & & & &   &  &  &  &  &  &  &  &   & & & & \\ \hline
\multicolumn{1}{l|}{HaRiM (BART$\_$large + cnn)} & \textbf{0.542}  & \multicolumn{1}{c|}{\textbf{0.581}}  & \textbf{0.620}  & \multicolumn{1}{c|}{\textbf{0.679}} & 0.336  & \multicolumn{1}{c|}{0.317}  & \textbf{0.463} & \multicolumn{1}{c|}{\textbf{0.437}} & 0.321  & \multicolumn{1}{c|}{0.268}  & 0.414  & \multicolumn{1}{c|}{0.391} & 0.167 & \multicolumn{1}{c|}{0.175} & 0.101 & 0.087 \\
\multicolumn{1}{l|}{HaRiM (BART-large + cnn + para)} & 0.515 & \multicolumn{1}{c|}{0.556} & 0.530 & \multicolumn{1}{c|}{0.610} & \textbf{0.387} & \multicolumn{1}{c|}{\textbf{0.356}} & 0.423  & \multicolumn{1}{c|}{0.408}  & \textbf{0.366} & \multicolumn{1}{c|}{\textbf{0.314}} & 0.426  & \multicolumn{1}{c|}{0.390} & 0.173 & \multicolumn{1}{c|}{0.172} & 0.037 & 0.042 \\ \hline
\end{tabular}%
}
\caption{\label{tab:spearman_pearson} Metric-Human correlation (segment-level) in Spearman's $r$ and Pearson's $\rho$. The best performance are bolded and second-bests are underlined.}
\end{table*}

\begin{table*}[]
\centering
\begin{tabular}{l|ll}
score                  & $r$             & $\rho$           \\ \hline
$\mathrm{log(H_{lm}/H_{s2s})}$          & 0.05          & 0.05          \\
$\mathrm{log(H_{lm}/H_{s2s})}~\_len$    & 0.05          & 0.05          \\
$\mathrm{H_{lm}/H_{s2s}}$               & 0.05          & 0.05          \\
$\mathrm{(H_{lm}/H_{s2s})}~\_len$        & 0.05          & 0.05          \\
$\mathrm{H_{s2s} * H_{lm}}$             & 0.23          & 0.10          \\
$\mathrm{(H_{s2s} * H_{lm})}~\_len$     & 0.00          & -0.01         \\
$\mathrm{log(H_{s2s} * H_{lm})}~\_len$  & 0.00          & 0.01          \\ \hline
$\mathrm{(H_{lm}-H_{s2s})}~\_len$       & 0.04          & 0.04          \\
$\mathrm{H_{lm}}$                    & 0.22          & 0.17          \\
$\mathrm{H_{lm}~\_len}$               & 0.04          & 0.02          \\
$\mathrm{H_{s2s}}$                   & 0.22          & 0.19          \\
$\mathrm{H_{s2s}~\_len}$             & -0.03         & -0.02         \\ \hline
$\mathrm{-HaRiM~\_lmless}$            & \textbf{0.46} & \textbf{0.50} \\
-HaRiM                    & \textbf{0.46} & \textbf{0.50} \\
-HaRiM (quintic) \_lmless & 0.45          & 0.40          \\
-HaRiM (quintic)          & 0.45          & 0.40          \\
-HaRiM\_top5mean          & 0.04          & 0.06          \\
-HaRiM\_bot5mean          & 0.14          & 0.17          \\ \hline
\end{tabular}%
\caption{Variation tested over FRANK \textit{CNN\/DailyMail} split. H denotes entropy. \textit{$\_$len} refers to length normalization. Entropy-based scores are performing worse. We also tested other variations for aggregating token-level scores into a scalar such as idf, tf-idf reweighting of HaRiM (not presented here) which do nothing more than worsening the correlation to human judgements similarly to top/bot 5 averaging. }
\label{tab:vary_marginf}
\end{table*}

\begin{table*}[ht]
\centering
\resizebox{\textwidth}{!}{
    \tabulinesep=0.8mm
    \begin{tabu}{lcccccccc}
    \multicolumn{1}{l|}{}                                        & \multicolumn{4}{c|}{\textbf{Abstractive}}                                                                                                & \multicolumn{4}{c}{\textbf{Extractive}}         \\
    \multicolumn{1}{l|}{Kendall's $\tau$}                         & \multicolumn{4}{c|}{1200 outputs}                                                                                                & \multicolumn{4}{c}{400 outputs}         \\                \hline
    \multicolumn{1}{l|}{\textbf{Metrics}}                        & \textbf{Con} & \textbf{Coh}   & \textbf{Flu}   & \multicolumn{1}{c|}{\textbf{Rel}}   & {\textbf{Con}}   & \textbf{Coh}  & \textbf{Flu} & \textbf{Rel} \\ \hline

    \textbf{N-gram matching}                                     &                     &                     &                &                &                &                                     &                     &                     \\ \hline
    \multicolumn{1}{l|}{ROUGE 1}                                 & 0.117               & 0.129              & 0.057          & \multicolumn{1}{c|}{0.219  }        & 0.094          & {0.209}          & 0.063               & 0.161               \\
    \multicolumn{1}{l|}{ROUGE 2}                                 & 0.107               & 0.128              & 0.041          & \multicolumn{1}{c|}{0.173     }     & 0.066          & {0.153}          & 0.030               & 0.118               \\
    \multicolumn{1}{l|}{ROUGE L}                                 & 0.114               & 0.096              & 0.071          & \multicolumn{1}{c|}{0.180   }       & 0.063          & {0.164}          & 0.033               & 0.123               \\
    \multicolumn{1}{l|}{METEOR}                                  & 0.094               & 0.091               & 0.025          & \multicolumn{1}{c|}{0.217     }     & 0.003          & {0.148}          & 0.121               & 0.201               \\
    \multicolumn{1}{l|}{sacreBLEU}                               & 0.109               & 0.201              & 0.103          & \multicolumn{1}{c|}{0.234   }       & 0.022          & {0.070}          & 0.091               & 0.147               \\ \hline
    
    \multicolumn{1}{l|}{ROUGE 1$\_$art}                            & 0.050               & -0.021               & -0.005          & \multicolumn{1}{c|}{0.114    }      & 0.104          & {0.117}          & 0.109              & 0.066              \\
    \multicolumn{1}{l|}{ROUGE 2$\_$art}                            & 0.150               & 0.020               & 0.073          & \multicolumn{1}{c|}{0.144   }       & 0.112          & {0.129}          & 0.113               & 0.077               \\
    \multicolumn{1}{l|}{ROUGE L$\_$art}                            & 0.157               & 0.045               & 0.083          & \multicolumn{1}{c|}{0.157 }         & {\ul 0.123}          & {0.166}          & 0.088               & 0.089              \\
    \multicolumn{1}{l|}{METEOR$\_$art}                             & 0.066               & -0.043               & 0.024          & \multicolumn{1}{c|}{0.082  }        & 0.107          & {0.087}          & 0.096               & 0.033              \\
    \multicolumn{1}{l|}{sacreBLEU$\_$art}                          & 0.023               & -0.016               & -0.036          & \multicolumn{1}{c|}{0.115 }         & 0.098          & 0.123         & 0.101               & 0.078              \\ \hline

    \textbf{N-gram stats}                                        &                     &                     &                &          &                &                                     &                     &                     \\ \hline
    \multicolumn{1}{l|}{NovelNgram$\_$4}                           & 0.241               & 0.230               & 0.245          & \multicolumn{1}{c|}{0.214}          & 0.042          & {0.085}          & 0.140               & 0.166               \\
    \multicolumn{1}{l|}{NovelNgram$\_$3}                           & 0.305               & 0.238               & {\ul 0.250}          & \multicolumn{1}{c|}{0.217 }         & 0.042          & {0.085}          & 0.147               & 0.170               \\
    \multicolumn{1}{l|}{NovelNgram$\_$2}                           & \textbf{0.315}               & 0.243               & 0.223          & \multicolumn{1}{c|}{0.218  }        & 0.045          & {0.084}          & 0.140               & 0.168         \\
    \multicolumn{1}{l|}{NovelNgram$\_$1}                           & 0.299               & 0.229               & 0.088          & \multicolumn{1}{c|}{0.189  }        & 0.040          & {0.088}          & 0.082               & 0.154               \\
    \multicolumn{1}{l|}{Length (no. tokens)}                     & -0.015              & -0.039              & -0.097         & \multicolumn{1}{c|}{0.120}          & 0.050          & {0.150}          & 0.068               & 0.137              \\ \hline

    \textbf{Contextual Embedding}                                &                     &                     &                &               &                &                                     &                     &                     \\ \hline
    \multicolumn{1}{l|}{BERTScore P}                             & 0.092               & 0.316              & 0.135          & \multicolumn{1}{c|}{0.229 }         & 0.043          & {0.019}           & 0.124               & 0.166               \\
    \multicolumn{1}{l|}{BERTScore R}                             & 0.124               & 0.257               & 0.071          & \multicolumn{1}{c|}{\textbf{0.309  }}        & 0.020          & 0.168            & 0.154               & \textbf{0.239}               \\
    \multicolumn{1}{l|}{BERTScore F1}                            & 0.110               & 0.330              & 0.124          & \multicolumn{1}{c|}{0.288  }        & 0.040          & {0.085}          & 0.154               & 0.229               \\ \hline
    \multicolumn{1}{l|}{BERTScore P$\_$art}                        & 0.263               & 0.334               & 0.225          & \multicolumn{1}{c|}{0.317  }        & 0.110          & {\ul 0.189}            & 0.187               & {\ul 0.234}              \\
    \multicolumn{1}{l|}{BERTScore R$\_$art}                        & 0.102               & 0.139               & 0.070         & \multicolumn{1}{c|}{ 0.239  }       &  0.083          & 0.141            & 0.141               & 0.160              \\
    \multicolumn{1}{l|}{BERTScore F1$\_$art}                       & 0.208               & 0.266               & 0.164          & \multicolumn{1}{c|}{0.319 }         & 0.112          & \textbf{0.196}          & 0.184               & 0.225              \\ \hline

    \textbf{Text Generation based}                               &                     &                     &                &             &                &                                     &                     &                     \\ \hline
    \multicolumn{1}{l|}{CBMI (BART$\_$base + cnn)*}                & 0.089               & -0.114               & -0.030          & \multicolumn{1}{c|}{0.016}        & 0.066          & 0.099          & -0.068             & 0.028              \\
    
    \multicolumn{1}{l|}{BARTScore (BART$\_$large + cnn)**}         & 0.222               & \textbf{0.368}               & 0.188          & \multicolumn{1}{c|}{0.288  }        & 0.099         & {0.102}          & \textbf{0.191}          & 0.178               \\
    \multicolumn{1}{l|}{BARTScore (BART$\_$large + cnn +   para)**}&  0.281              & 0.350               & 0.249          & \multicolumn{1}{c|}{0.303  }        & 0.128          & {0.111}          & {\ul 0.188}         & 0.180               \\ \hline
    \textbf{Proposed}                                            &                     &                     &                &                                     &                &               &                     &                    \\ \hline
    \multicolumn{1}{l|}{\textbf{$\mathrm{HaRiM}^+$ (BART$\_$large + cnn)}}& 0.278           & {\ul 0.366}               & 0.219          & \multicolumn{1}{c|}{{\ul 0.308}}          & 0.098         & {0.120}        & 0.185               & 0.190               \\
    \multicolumn{1}{l|}{$\mathrm{HaRiM}^+$ (BART$\_$large + cnn + para)} & {\ul 0.306}            & 0.339               & \textbf{0.260}          & \multicolumn{1}{c|}{0.306 }         & \textbf{0.126}         & {0.110}        & 0.176               & 0.183               \\ \hline

    \end{tabu} 
}
\caption{\label{tab:absextseparate}Metric-to-human judgement correlation (segment-level) reported in Kendall's $\tau$. \textbf{Bold}-face values are the largest correlating metrics, underlined are second-large values amongst the metrics. Hallucination Risk($\mathrm{HaRiM}^+$) outperforms others in most criteria. We provide permutation test result in Appendix. *\cite{wu-etal-2021-conditional}, **\cite{NEURIPS2021_e4d2b6e6}}
\end{table*}

\begin{figure*}
\centering
\includegraphics[width=\textwidth]{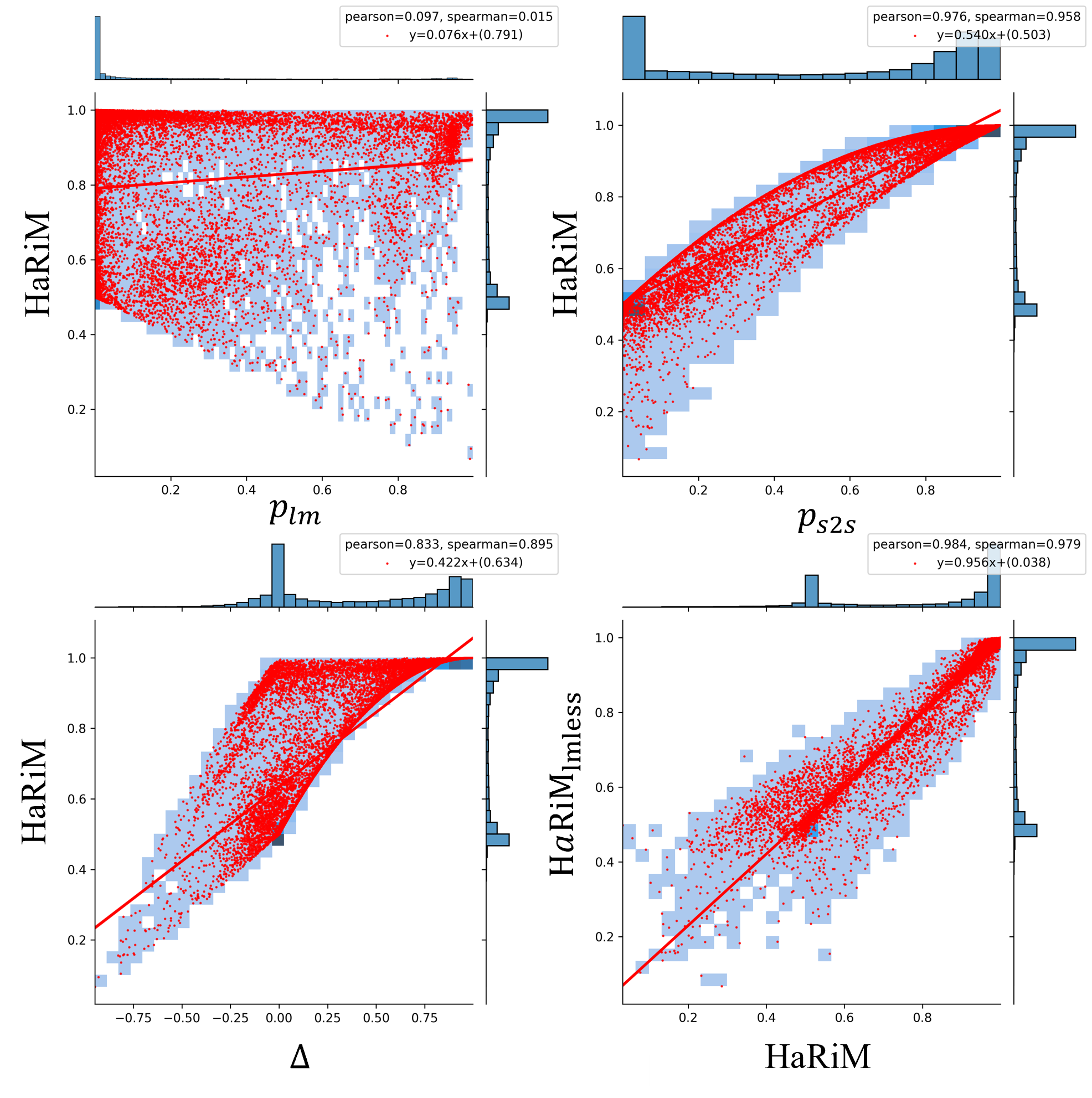} 
\caption{\label{fig:HaRiMvVars}Effect of each variable to HaRiM. $\Delta$ represents $p_{s2s}-p_{lm}$. The last figure at the righter down shows the effect of replacing auxiliary LM probability with empty-sourced decoder inference ($HaRiM_{lmless}$). Figure \ref{fig1:replacelm} shows article-summary pair as a datapoint in the plot, here we show each token of the decoded output as a datapoint.}
\end{figure*}
\subsection{Score Scales: HaRiM$^+$, HaRiM, and Log-likelihood }
\begin{figure*}
\centering
\includegraphics[width=\textwidth]{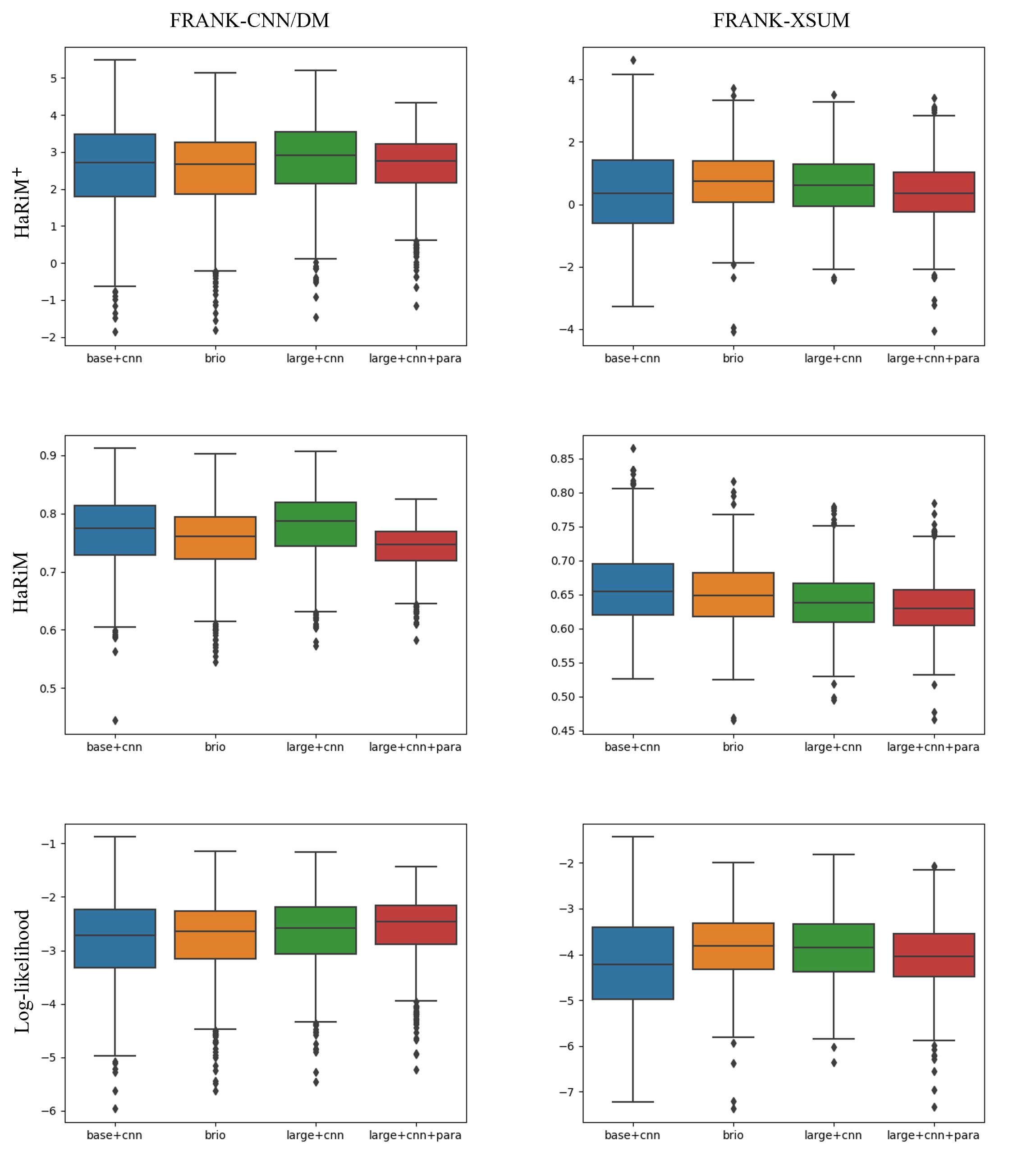} 
\cprotect\caption{\label{fig:scorescales} Boxplot of HaRiM and log-likelihood scales, varying with the evaluating summarizer weight. $\verb|base+cnn|$: BART-base fine-tuned on \textit{CNN/DailyMail}, $\verb|brio|$: BRIO \cite{meng-etal-2021-bringing}, $\verb|large+cnn|$: BART-large fine-tuned on \textit{CNN/DailyMail}, $\verb|large+cnn+para|$: further fine-tuned checkpoint of the previous model on ParaBank2 corpus as suggested in \cite{NEURIPS2021_e4d2b6e6}. }
\end{figure*}

\begin{figure*}
\centering
\includegraphics[width=\columnwidth]{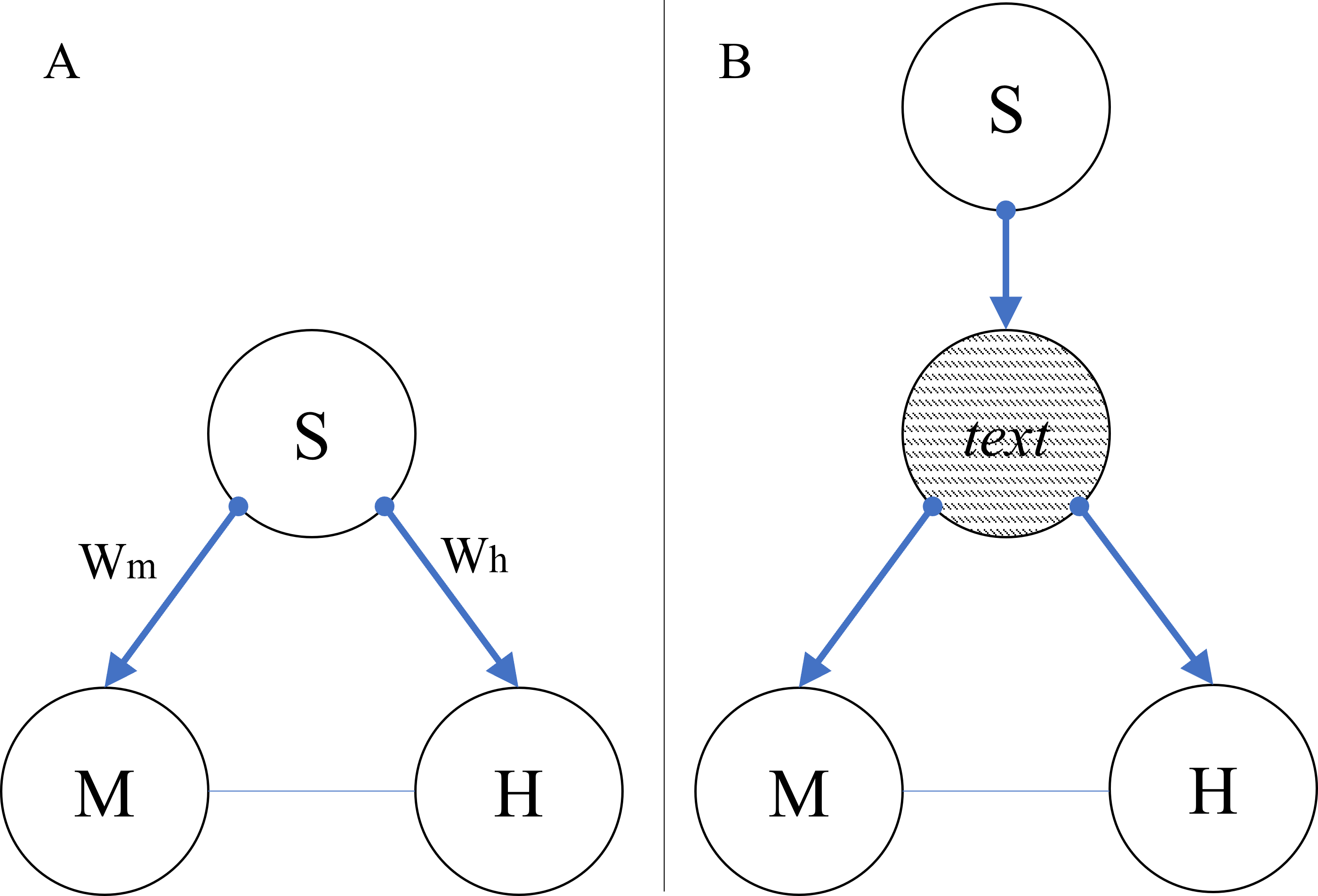}
\caption{\label{fig:graphicalmodel} Graphical model representation attributing to the factors that affects metric ($M$)-human ($H$) correlation. A is the graphical model that supports the use of partial correlation as argued in \cite{pagnoni-etal-2021-understanding}. B is the graphical model that adheres to our argument that why should we measure correlation, ignoring the effect of the generation system ($S$) whose effect is hindered by observed child node, \textbf{$text$}.}
\end{figure*}

\begin{figure}
\centering
\includegraphics[width=\columnwidth]{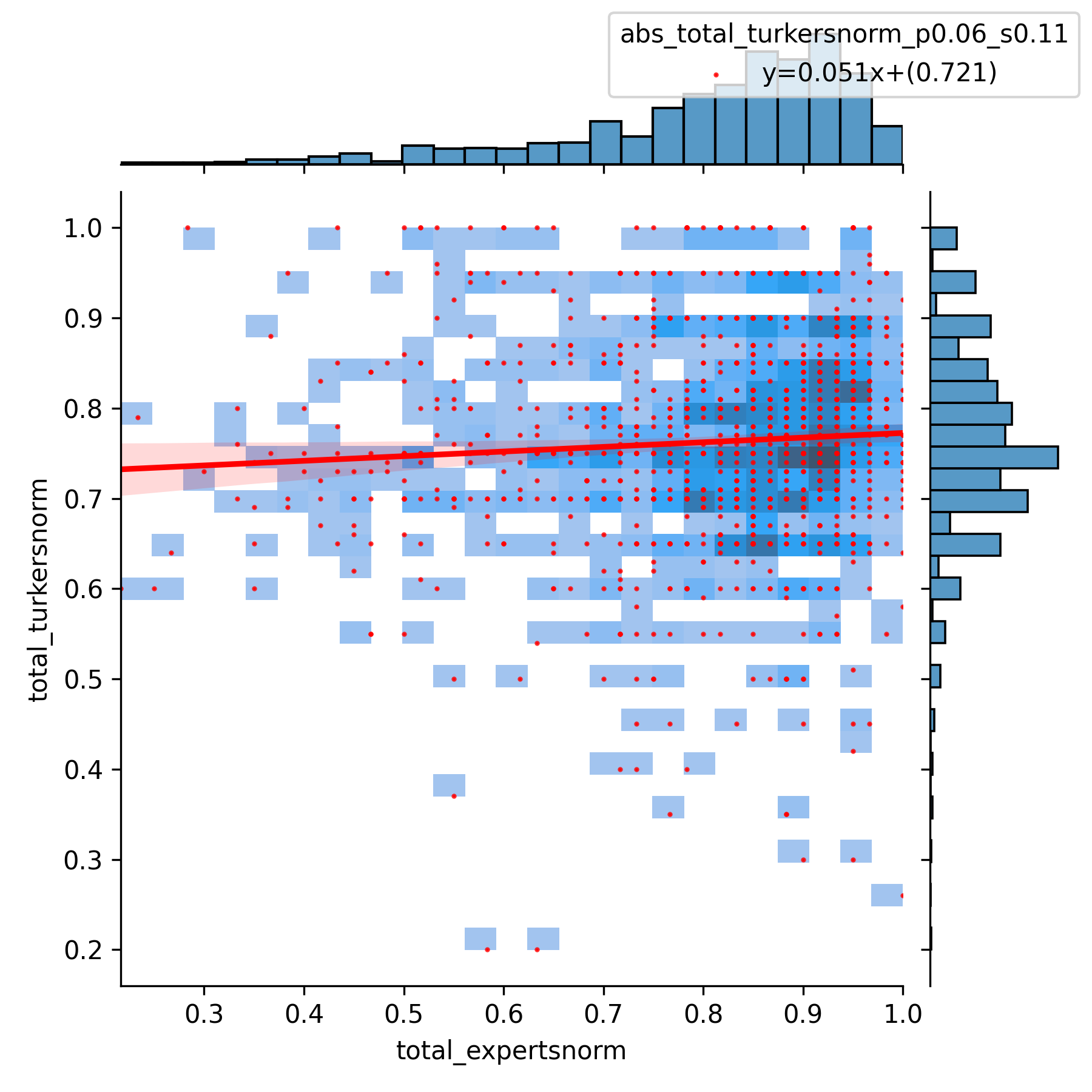}
\caption{\label{fig:spurrious_abs} Averaged experts' judgements vs. Averged turkers' judgements on SummEval, (datapoints are outputs from \textbf{abstractive} summarization models) }
\end{figure}

\begin{figure}
\centering
\includegraphics[width=\columnwidth]{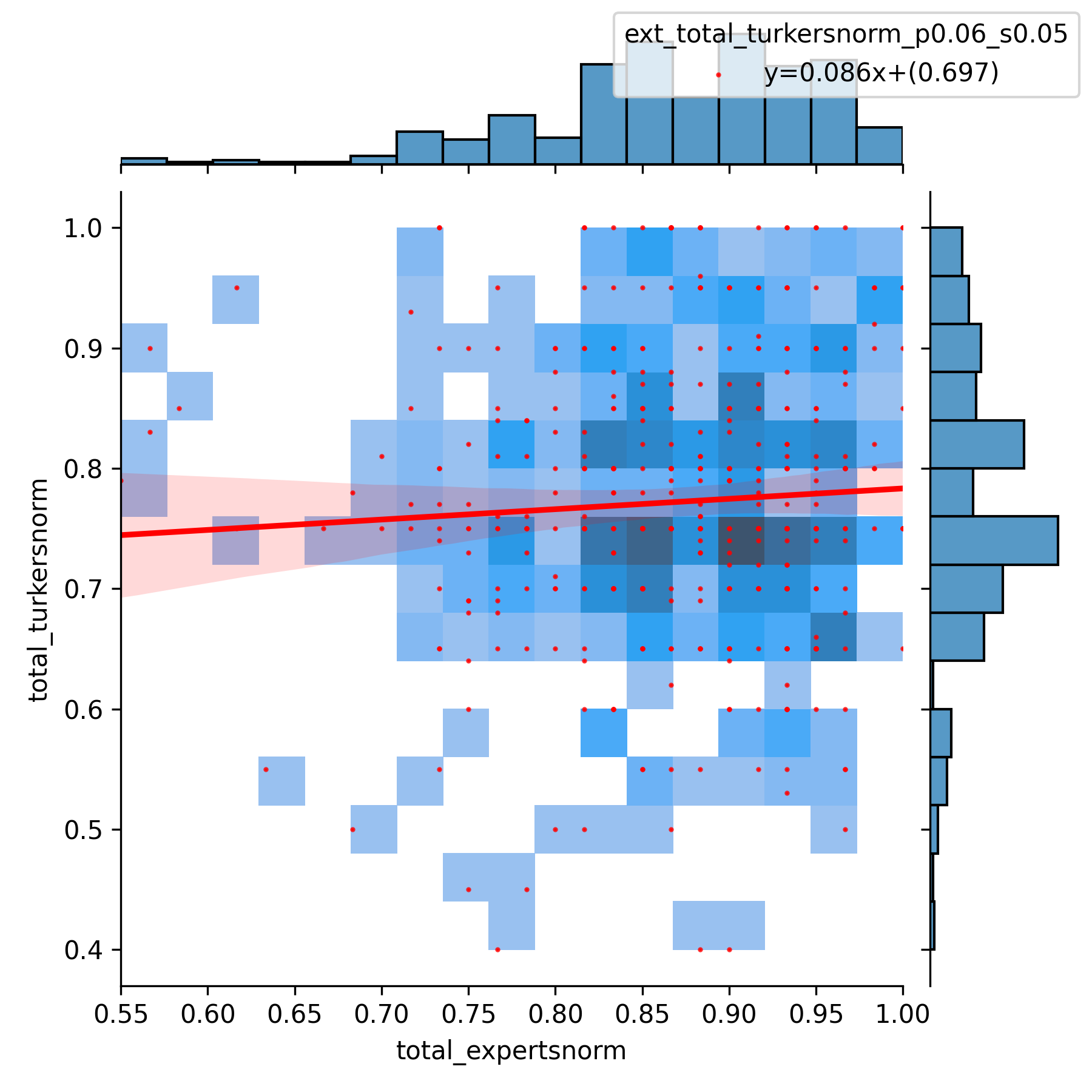}
\caption{\label{fig:spurrious_ext} Averaged experts' judgements vs. Averged turkers' judgements on SummEval, (datapoints are outputs from \textbf{extractive} summarization models)}
\end{figure}

In Figure \ref{fig:scorescales}, we visualize score scales of proposed HaRiM$^+$, HaRiM, and log-likelihood varying summarization model checkpoints.
We considered scale of each HaRiM and loglikelihood to decide the mixing coefficient $\lambda$ (searched over 0.1, 1, 5, 7, 8, 10, 20 and finally chose 7 to use). 

\clearpage

\clearpage

\begin{table*}[ht]
\centering
\resizebox{\textwidth}{!}{
\tabulinesep=0.6mm
\begin{tabu}{@{}clcc@{}}
\toprule
\multicolumn{4}{c}{\textbf{Source Article}} \\ \midrule
\multicolumn{4}{m{22.6cm}}{A youngster has emulated Barcelona star Martin Montoya and scored an audacious 27-yard goal into a basketball hoop - twice. Schoolboy Frankie Franz watched the Spanish right-back pull off the staggering trick shot in a video recorded at Barcelona’s Ciutat Esportiva training ground earlier in the month. The viral clip shows the 23-year-old defender lifting the ball into the net to the sound of gasps from his team mates at the Catalonia club. Joking  that he could do the same with his mum and grandmother, nine-year-old Frankie, who is an academy player with Dagenham and Redbridge Football Club, took to the garden to have a go. He moved the basketball hoop into the middle of the goal and after a little run up sent the ball straight through the net first time. In the video he can be seen turning to face the camera looking absolutely gob-smacked as he places his hands behind his head. A slow motion look at the clip captures the trajectory of the ball, which bounces off the backboard and goes through the hoop below. His mother Lucy, 32, of Upminster, east London, said: ‘He loves football and after he saw the Barcelona player do the show he said ‘I’ll be able to do that’. (...) The talented young centre midfielder has played football since he could walk and dreams of one day turning out for Real Madrid or Barcelona. The  young centre midfielder is an academy player with Dagenham and Redbridge Football Club and dreams of one day turning out for Real Madrid or Barcelona.} \\ \midrule
\multicolumn{1}{c|}{\textbf{Model}} & \multicolumn{1}{c|}{\textbf{Summary}} & \multicolumn{1}{c|}{\textbf{$\mathrm{HaRiM}^+$ Score $\uparrow$}} & \textbf{Score Gain $\uparrow$} \\ \midrule
\multicolumn{1}{c|}{\textbf{Reference}} & \multicolumn{1}{m{12cm}|}{frankie franz watched the right-back pull off the audacious shot in a video . nine-year-old joked with his mum and grandmother that he could make it . youngster moved hoop into middle of the garden and twice achieved feat . frankie is an academy player with dagenham and redbridge football club . he plays centre midfield and dreams of one day turning out for barcelona .} & \multicolumn{1}{c|}{3.7448} & - \\ \midrule
\multicolumn{1}{c|}{\textbf{\begin{tabular}[c]{@{}c@{}}Self-generation\\ (BART-large+cnn)\end{tabular}}} & \multicolumn{1}{m{12cm}|}{frankie franz watched the spanish right-back pull off the staggering trick shot in a video recorded at barcelona’s ciutat esportiva training ground earlier in the month. the viral clip shows the 23-year-old defender lifting the ball into the net to the sound of gasps from his team mates at the catalonia club. joking that he could do the same with his mum and grandmother, frankie took to the garden to have a go. he moved the basketball hoop into the middle of the goal and after a little run up sent the ball straight through the net first time.} & \multicolumn{1}{c|}{4.5869} & +0.8421 \\ \midrule
\multicolumn{1}{c|}{\textbf{\begin{tabular}[c]{@{}c@{}}BottomUpSummary\\ (Factuality=\textcolor{red}{0.0})\end{tabular}}} & \multicolumn{1}{m{12cm}|}{frankie franz watched the spanish right-back pull off the trick shot in a video recorded at barcelona 's catalonia club . the 23-year-old defender took to the garden to have a go and moved the basketball hoop into the net to the goal . his mother lucy , 32 , said : ` me said ` i will be able to do . ' .} & \multicolumn{1}{c|}{2.6875} & -1.0573 \\ \midrule
\multicolumn{1}{c|}{\textbf{\begin{tabular}[c]{@{}c@{}}Reference\\ (w/ wrong subject)\end{tabular}}} & \multicolumn{1}{m{12cm}|}{\textcolor{red}{martin montoya} watched the right-back pull off the audacious shot in a video . nine-year-old joked with his mum and grandmother that he could make it . youngster moved hoop into middle of the garden and twice achieved feat . frankie is an academy player with dagenham and redbridge football club . he plays centre midfield and dreams of one day turning out for barcelona .} & \multicolumn{1}{c|}{3.7903} & +0.0455 \\ \midrule
\multicolumn{1}{c|}{\textbf{\begin{tabular}[c]{@{}c@{}}Reference\\ (w/ negation)\end{tabular}}} & \multicolumn{1}{m{12cm}|}{frankie franz \textcolor{red}{did not} watch the right-back pull off the audacious shot in a video . nine-year-old joked with his mum and grandmother that he could make it . youngster moved hoop into middle of the garden and twice achieved feat . frankie is an academy player with dagenham and redbridge football club . he plays centre midfield and dreams of one day turning out for barcelona .} & \multicolumn{1}{c|}{3.3877} & -0.3571  \\ \bottomrule

\end{tabu}}
\caption{\label{tab:qual_neg1}Testing $\mathrm{HaRiM}^+$ metric under hallucination detecting scenario. Part of the source article, which is irrelevant to the summaries are omitted for clarity. The words highlighted red are hallucinated information deliberately injected to the reference. BottomUpSummary refers to abstractive summarization system suggested in \cite{gehrmann-etal-2018-bottom}.}
\end{table*}

\begin{table*}[ht]
\centering
\resizebox{\textwidth}{!}{
\tabulinesep=0.6mm
\begin{tabu}{@{}clcc@{}}
\toprule
\multicolumn{4}{c}{\textbf{Source Article}} \\ \midrule
\multicolumn{4}{m{22.6cm}}{Spain's 2-0 defeat by Holland on Tuesday brought back bitter memories of their disastrous 2014 World Cup, but coach Vicente del Bosque will not be too worried about a third straight friendly defeat, insists Gerard Pique. Holland, whose 5-1 drubbing of Spain in the group stage in Brazil last year marked the end of the Iberian nation's six-year domination of the world game, scored two early goals at the Amsterdam Arena and held on against some determined Spain pressure in the second half for a 2-0 success. (...) Stefan de Vrij (right) headed Holland in front against Spain at the Amsterdam Arena on Tuesday Gerard Pique (left) could do nothing to stop Davy Klaassen doubling the Dutch advantage Malaga forward Juanmi and Sevilla midfielder Vitolo became the 55th and 56th players to debut under Del Bosque, (...) `The national team's state of health is good,' centre back Gerard Pique told reporters. 'We are in a process where players are coming into the team and gathering experience,' added the Barcelona defender. `We are second in qualifying (for Euro 2016) and these friendly games are for experimenting. (...)} \\ \midrule 

\multicolumn{1}{c|}{\textbf{Model}} & \multicolumn{1}{c|}{\textbf{Summary}} & \multicolumn{1}{c|}{\textbf{HaRiM${^+}$ Score} $\uparrow$} & \textbf{Score Gain $\uparrow$} \\ \midrule
\multicolumn{1}{c|}{\textbf{Reference}} & \multicolumn{1}{m{13.2cm}|}{holland beat spain 2-0 at the amsterdam arena on tuesday night . stefan de vrij and davy klaassen scored goals for holland . defeat recalls horror 5-1 defeat by holland at the world cup . vicente del bosque used game to give younger spain players a chance .} & \multicolumn{1}{c|}{2.7096} & - \\ \midrule
\multicolumn{1}{c|}{\textbf{\begin{tabular}[c]{@{}c@{}}Self-generation\\ (BART-large+cnn)\end{tabular}}} & \multicolumn{1}{m{13.2cm}|}{holland's 5-1 drubbing of spain last year marked the end of the iberian nation's six-year domination of the world game. spain's 2-0 defeat by holland on tuesday brought back bitter memories of their disastrous 2014 world cup, but coach vicente del bosque will not be too worried about a third straight friendly victory. 'the national team's state of health is good,' says defender gerard pique} & \multicolumn{1}{c|}{3.7338} & +1.0242 \\ \midrule
\multicolumn{1}{c|}{\textbf{\begin{tabular}[c]{@{}c@{}}RNN-S2S\\ (Factuality=\textcolor{red}{0.0})\end{tabular}}} & \multicolumn{1}{m{13.2cm}|}{holland beat spain 2-0 in the group stage in brazil on tuesday night . del bosque will be hoping to find the right mix of players to the world cup . gerard pique could make the right mix of players to the tournament .} & \multicolumn{1}{c|}{2.669} & -1.5074 \\ \midrule
\multicolumn{1}{c|}{\textbf{\begin{tabular}[c]{@{}c@{}}Reference\\ (w/ wrong subject)\end{tabular}}} & \multicolumn{1}{m{13.2cm}|}{\textcolor{red}{del bosque} beat spain 2-0 at the amsterdam arena on tuesday night . stefan de vrij and davy klaassen scored goals for holland . defeat recalls horror 5-1 defeat by holland at the world cup . vicente del bosque used game to give younger spain players a chance .} & \multicolumn{1}{c|}{2.4039} & -0.3057 \\ \midrule
\multicolumn{1}{c|}{\textbf{\begin{tabular}[c]{@{}c@{}}Reference\\ (w/ negation)\end{tabular}}} & \multicolumn{1}{m{13.2cm}|}{holland \textcolor{red}{could not} beat spain 2-0 at the amsterdam arena on tuesday night . stefan de vrij and davy klaassen scored goals for holland . defeat recalls horror 5-1 defeat by holland at the world cup . vicente del bosque used game to give younger spain players a chance .} & \multicolumn{1}{c|}{2.3759} & -0.3337 \\ \bottomrule

\end{tabu}}
\caption{\label{tab:quala}Testing HaRiM${^+}$ metric under hallucination detecting scenario. Part of the source article, which is irrelevant to the summaries are omitted for clarity. The words highlighted red are hallucinated information deliberately injected to the reference. RNN-S2S refers to \cite{NIPS2014_a14ac55a}.}
\end{table*}

\begin{table*}[ht]
\centering
\resizebox{\textwidth}{!}{
\tabulinesep=0.6mm
\begin{tabu}{@{}clcc@{}}
\toprule
\multicolumn{4}{c}{\textbf{Source Article}} \\ \midrule
\multicolumn{4}{m{22.6cm}}{(CNN)Soon, America will be too fat to fight. Forget about rampant diabetes, heart attacks and joint problems -- the scariest consequence arising out of our losing battle with the bulge is the safety of our country. In about five years, so many young Americans will be grossly overweight that the military will be unable to recruit enough qualified soldiers. That alarming forecast comes from Maj, Gen. Allen Batschelet, who is in charge of U.S. Army Recruiting Command. Obesity, he told me, ``is becoming a national security issue." I was so taken aback by Batschelet's statement that I felt the need to press him. Come on! Obesity? A national security crisis? The General didn't blink. ``In my view, yes." Of the 195,000 young men and women who signed up to fight for our country, only 72,000 qualified. Some didn't make the cut because they had a criminal background, or a lack of education, or too many tattoos. But a full 10\% didn't qualify because they were overweight. Before you accuse me of sensationalizing, it's that 10\% figure that worries General Batschelet the most. ``The obesity issue is the most troubling because the trend is going in the wrong direction," he said. ``We think by 2020 it could be as high as 50\%, which mean only 2 in 10 would qualify to join the Army." He paused. ``It's a sad testament to who we are as a society right now." The problem is so worrisome for the Army that recruiters have become fitness coaches, like the trainers on the NBC show, ``The Biggest Loser." Yes, your tax dollars pay for Army recruiters to play Dolvett Quince or Jillian Michaels to whip could-be recruits into shape with the hope they can diet and exercise their way to become real recruits. If they lose enough weight, they're sent to boot camp. Some make it; many don't. But, General Batschelet told me the Army must try. ``We are the premier leader on personal development in the world," he told me. ``We want to see you grow and become a leader. That is a great strength in our Army." Except the Army never considered the type of growth it's now contending with. Nowadays ``personal development" means working on both character and ... girth. The general, along with so many others in this country, is struggling with why so many Americans, despite all the warnings, continue to eat too much and exercise too little. I have a theory. It ain't pretty. But it's got to be true: We just don't care. ``The acceptance of obesity is prevalent," according to Claire Putnam, an obstetrician and gynecologist who believes obesity is a national crisis right now. ``When you look around you, 70\% of adults are overweight or obese. It's seems normal," she said. Just look at the numbers: More than one-third of U.S. adults are obese. Seventeen percent of all children and adolescents in the U.S. are obese. That's triple the rate from just a generation ago. So, maybe we should face the fact that we've grown comfortable with our girth. It is crystal clear we haven't the foggiest idea of who needs to lose weight and who doesn't. Just the other day, Twitter trolls scolded the singer, Pink, for gaining weight. Pink is not remotely fat. Neither is Selena Gomez, haters. Or Britney Spears, hecklers. If 70\% of us are overweight in this country, why are there so many willing to fat-shame people who are not remotely obese? Maybe it's easier to criticize others for carrying extra weight than to admit we have a weight problem ourselves. Because it is abundantly clear we are wallowing in denial. Dr. Putnam points to one of Kaiser Permanante's medical questionnaires. You know, the paperwork patients are asked to fill out before they see the doctor. There is actually a box on the form that allows the patient to ``opt out of talking about obesity." Some patients refuse to step on the scale. ``You want to be sensitive to that patient," Putnam told me. ``You don't want to nag. But, doctors need to step in and say we need to fix this." CNN's chief medical correspondent, Dr. Sanjay Gupta, agrees with Putnam. ``Perceptions of weight are a big part of the problem," he said to me. ``If a person is overweight -- as difficult as it is -- they ought to be told. You know, this issue reminds me of the issue with concussions. We should call them what they really are: a brain injury, not 'getting your bell rung.' In the same vein, we should tell people who are overweight or obese that, clinically, they're 'overweight' or 'obese' and at risk for just about every chronic disease in the book." In other words, chubby is not the proper way to describe a person who is obese. Just like ``fat" is not the proper term for Pink or Selena Gomez. And, yes, semantics matter. According to the CDC, 81\% of overweight boys and 71\% of overweight girls believe they are just the right weight. We've clearly lost our perspective on what's normal when it comes to a healthy weight. So much so it's becoming a national security problem. So what will it take? The answer cannot be the U.S Army.} \\ \midrule
\multicolumn{1}{c|}{\textbf{Model}} & \multicolumn{1}{c|}{\textbf{Summary}} & \multicolumn{1}{c|}{\textbf{$\mathrm{HaRiM}^+$ Score $\uparrow$}} & \textbf{Score Gain $\uparrow$} \\ \midrule
\multicolumn{1}{c|}{\textbf{Reference}} & \multicolumn{1}{m{13.2cm}|}{in a few years , the military will be unable to recruit enough qualified soldiers because of america 's obesity problem . carol costello : we have a serious national security issue at hand , but it 's within our control if we could own up to it .} 
& \multicolumn{1}{c|}{1.6283} 
& - \\ \midrule

\multicolumn{1}{c|}{\textbf{\begin{tabular}[c]{@{}c@{}}Self-generation\\ (BART-large+cnn)\end{tabular}}} & \multicolumn{1}{m{13.2cm}|}{of the 195,000 young men and women who signed up to fight for our country, only 72,000 qualified. a full 10\% didn't qualify because they were overweight. ``it's a sad testament to who we are as a society right now," says maj, gen. allen batschelet.} 
& \multicolumn{1}{c|}{5.061} 
& +3.4327 \\ \midrule

\multicolumn{1}{c|}{\textbf{\begin{tabular}[c]{@{}c@{}}BottomUpSummary\\ (Factuality=\textcolor{red}{0.0})\end{tabular}}} & \multicolumn{1}{m{13.2cm}|}{many young americans will be overweight that the military will be able to recruit enough soldiers . gen. allen batschelet is a national security issue for the u.s. army . he says the obesity issue is so many that it 's too fat to fight .} 
& \multicolumn{1}{c|}{1.3449} 
& -0.2834  \\ \midrule

\multicolumn{1}{c|}{\textbf{\begin{tabular}[c]{@{}c@{}}Reference\\ (w/ wrong subject)\end{tabular}}} & \multicolumn{1}{m{13.2cm}|}{in a few years , the military will be unable to recruit enough qualified soldiers because of america 's obesity problem . \textcolor{red}{claire putnam} : we have a serious national security issue at hand , but it 's within our control if we could own up to it .} 
& \multicolumn{1}{c|}{1.5808} 
& -0.0475 \\ \midrule

\multicolumn{1}{c|}{\textbf{\begin{tabular}[c]{@{}c@{}}Reference\\ (w/ negation)\end{tabular}}} & \multicolumn{1}{m{13.2cm}|}{in a few years , the military will be unable to recruit enough qualified soldiers because of america 's obesity problem . carol costello : we \textcolor{red}{do not} have a serious national security issue at hand , but it 's within our control if we could own up to it .} 
& \multicolumn{1}{c|}{1.4759} 
& -0.1524 \\ \bottomrule

\end{tabu}}
\caption{\label{tab:qual_neg2}Testing $\mathrm{HaRiM}^+$ metric under hallucination detecting scenario. Part of the source article irrelevant to the summaries are omitted for clarity. The words highlighted red are hallucinated information deliberately injected to the reference. BottomUpSummary refers to abstractive summarization system suggested in \cite{gehrmann-etal-2018-bottom}.}
\end{table*}

\begin{table*}[ht]
\centering
\resizebox{\textwidth}{!}{
\tabulinesep=0.6mm
\begin{tabu}{@{}clcc@{}}
\toprule
\multicolumn{4}{c}{\textbf{Source Article}} \\ \midrule
\multicolumn{4}{m{22.6cm}}{It's well known that exercise can make your muscles bigger. Now, a study has found it may make your brain larger, too. Physical activity can increase grey matter in the brain, increasing the size of areas that contribute to balance and coordination, according to  Health Day news. The changes in the brain may have health implications in the long-term, such as reducing the risk of falling, said the study's author, Dr Urho Kujala, of the University of Jyvaskyla. Scroll down for video Exercise can increase the size of areas of the brain that contribute to balance and coordination, a study found It could also reduce the risk of being immobile in older age, he added. Dr Kujala said physical activity has already been linked to a number of health benefits, such as lower levels of body fat, reduced heart disease risk factors, better memory and thinking, and a lower risk of type 2 diabetes. But he and his team wanted to understand how exercise affects the brain. They recruited 10 pairs of identical twins, who were all men aged 32 to 36 years. Focusing on twins, who have the same DNA, would allow researchers to see how their environment affects their bodies. In each pair of twins, one brother had exercised more over the past three years than the other, though they reported they carried out similar levels of exercise earlier in their lives. Dr Kujala said: 'On average, the more active members of twin pairs were jogging three hours more per week compared to their inactive co-twins.' The twins had MRI scans of their brains so researchers could see whether physical activity had any impact on the size of their brains, and specific regions. Exercise didn't seem to affect the size of the brain as a whole, Dr Kujala said. But there was a connection between more activity and more brain volume in areas related to movement, he added. Previous research found exercise is linked to lower levels of body fat, a reduced risk of heart disease, better memory and thinking, and a lower risk of type 2 diabetes The twins who exercised more did a better job of controlling their blood sugar, which reduces the risk of diabetes, a finding which is already well-known. The study was published in the journal Medicine \& Science in Sports \& Exercise. It comes after US researchers found regular exercise can also make you smarter. University of South Carolina experts found  regular treadmill sessions create more mitochondria -  structures in the cells that produce the body's energy - in the brain. This energy boost helped the brain to work faster and more efficiently, effectively keeping it younger, researchers said. In the short term this could reduce mental fatigue and sharpen your thinking in between gym sessions. And building up a large reservoir of mitochondria in the brain could also create a 'buffer' against age-related brain diseases such as Alzheimer's.} \\ \midrule
\multicolumn{1}{c|}{\textbf{Model}} & \multicolumn{1}{c|}{\textbf{Summary}} & \multicolumn{1}{c|}{\textbf{$\mathrm{HaRiM}^+$ Score $\uparrow$}} & \textbf{Score Gain $\uparrow$} \\ \midrule
\multicolumn{1}{c|}{\textbf{Reference}} & \multicolumn{1}{m{13.2cm}|}{study : exercising increases the amount of grey matter in the brain . it makes areas of the brain that control balance and co-ordination bigger . in the long term this could reduce the risk of falling or becoming immobile . previous studies show exercise can stave off alzheimer 's and diabetes .} 
& \multicolumn{1}{c|}{3.5959} 
& - \\ \midrule

\multicolumn{1}{c|}{\textbf{\begin{tabular}[c]{@{}c@{}}Self-generation\\ (BART-large+cnn)\end{tabular}}} & \multicolumn{1}{m{13.2cm}|}{physical activity can increase grey matter in the brain, a study found. it can increase the size of areas that contribute to balance and coordination. changes may have health implications in the long-term, such as reducing the risk of falling, said the study's author, dr urho kujala, of the university of jyvaskyla.} 
& \multicolumn{1}{c|}{5.8816} 
& +2.2857 \\ \midrule

\multicolumn{1}{c|}{\textbf{\begin{tabular}[c]{@{}c@{}}BERTSum\\ (Factuality=1.0)\end{tabular}}} & \multicolumn{1}{m{13.2cm}|}{exercise can increase grey matter in the brain , increasing the size of areas that contribute to balance and coordination . study 's author , dr urho kujala , of the university of jyvaskyla , said physical activity has already been linked to a number of health benefits , such as lower levels of body fat , reduced heart disease risk factors , better memory and thinking , and a lower risk of type 2 diabetes .} 
& \multicolumn{1}{c|}{3.8014} 
& +0.2055  \\ \midrule

\multicolumn{1}{c|}{\textbf{\begin{tabular}[c]{@{}c@{}}Reference\\ (w/ wrong subject)\end{tabular}}} & \multicolumn{1}{m{13.2cm}|}{study : exercising increases the amount of \textcolor{red}{mitochondria} in the brain . it makes areas of the brain that control balance and co-ordination bigger . in the long term this could reduce the risk of falling or becoming immobile . previous studies show exercise can stave off alzheimer 's and diabetes .} 
& \multicolumn{1}{c|}{3.3481} 
& -0.2478 \\ \midrule

\multicolumn{1}{c|}{\textbf{\begin{tabular}[c]{@{}c@{}}Reference\\ (w/ negation)\end{tabular}}} & \multicolumn{1}{m{13.2cm}|}{study : exercising \textcolor{red}{does not} increase the amount of grey matter in the brain . it makes areas of the brain that control balance and co-ordination bigger . in the long term this could reduce the risk of falling or becoming immobile . previous studies show exercise can stave off alzheimer 's and diabetes .} 
& \multicolumn{1}{c|}{3.2656} 
& -0.3303 \\ \bottomrule

\end{tabu}}
\caption{\label{tab:qual_pos1}Testing $\mathrm{HaRiM}^+$ metric under hallucination detecting scenario. Part of the source article irrelevant to the summaries are omitted for clarity. The words highlighted red are hallucinated information deliberately injected to the reference. BERTSum refers to extractive summarization system suggested in \cite{liu-lapata-2019-text}.}
\end{table*}

\begin{table*}[ht]
\centering
\resizebox{\textwidth}{!}{
\tabulinesep=0.6mm
\begin{tabu}{@{}clcc@{}}
\toprule
\multicolumn{4}{c}{\textbf{Source Article}} \\ \midrule
\multicolumn{4}{m{22.6cm}}{The respected law professor from Philadelphia now being investigated after allegedly emailing students a link to  pornographic  footage, was once a contestant on Who Wants to Be a Millionaire, it has emerged. Lisa McElroy, a 50-year-old Drexel professor, appeared on the show in 2010 while it was still hosted my Meredith Vieira. And like her apparent March 31 email mishap, her game show appearance ended with a very public mistake. McElroy, who teaches legal writing, got tripped up on the \$12,500 level after flying through the first few questions, notes Philly.com. Wishes she was a millionaire: Drexel law profesor professor Lisa McElroy allegedly sent a link to a pornographic website to her students. In 2010, she appeared on the TV game show Who Wants to Be a Milionaire Mother of two: The mother of two shared an anecdote with then-host Meredith Vieira about having to scramble to find a babysitter for her kids and someone to teach her class after learning she was to appear on the show just two days before taping Lost it: McElroy was tripped up on the \$12,500 question. Despite having used two lifelines, she answered wrong and walked away with around \$5,000 The questions read: 'As a result of General Motor’s bankruptcy declaration in 2009, what foreign government became one of its largest shareholders?' Even after using two of her lifelines to narrow down the answer, McElroy answered China, which was incorrect. The correct answer was Canada. She walked away with around \$5,000. McElroy, who is a children's book and biography author, is apparently also a mother. She opened the appearance by sharing an anecdote with Vieira about having to scramble to find a babysitter after being informed she was chosen to be on Millionaire jsut two days prior to taping. She's accused of sending the inappropriate message this past March 31 under the subject line: 'Great article on writing briefs.' However, when recipients opened the enclosed link, philly.com  reports that they were directed to a video of 'a woman engaging in a sexually explicit act'. Lisa McElroy, 50, who teaches legal writing at Drexel University, reportedly sent the inappropriate message on March 31 baring the subject line: 'Great article on writing briefs' Following a number of complaints, the college issued an apology to students. The message read: 'As you may be aware, some students erroneously received an email this morning directing them to a... post that included some inappropriate material. 'We take this matter seriously and apologize for any upset it may have caused.' The university says federal law requires it investigate all reports of inappropriate behaviors of a sexual nature. McElroy did not immediately respond to an email sent to her university account by the Associated Press. When recipients opened the enclosed link, philly.com reports that they were directed to a video of 'a woman engaging in a sexually explicit act' It's not the first time the married mother-of-two has appeared in the spotlight. She is also an accomplished author with a number of published biographies and children's books. On her website, www.lisamcelroy.com, she describes herself as a 'Supreme Court junkie.' She adds that her favorites ways of relaxing include 'crawling under the covers with a dog or two and a really good book' or 'hanging out' with her two adolescent daughters. Regarding the recent email scandal, David Lat - a lawyer and legal commenter -suggests she could have been 'hacked' or made a 'copy/paste error'. While an internal investigation gets underway, it's been reported that McElroy has been placed on administrative leave. While an internal investigation gets underway, it's been reported that McElroy has been placed on administrative leave from Drexel University (seen above)} \\ \midrule
\multicolumn{1}{c|}{\textbf{Model}} & \multicolumn{1}{c|}{\textbf{Summary}} & \multicolumn{1}{c|}{\textbf{$\mathrm{HaRiM}^+$ Score $\uparrow$}} & \textbf{Score Gain $\uparrow$} \\ \midrule
\multicolumn{1}{c|}{\textbf{Reference}} & \multicolumn{1}{m{13.2cm}|}{lisa mcelroy , 50 , who teaches legal writing at drexel university , reportedly sent the ` inappropriate ' message on march 31 . when recipients clicked the enclosed link , they were allegedly directed to a video of ' a woman engaging in a sexually explicit act ' . mcelroy appeared on the popular game show in 2010 with then-host meredith vieira but lost the game after reaching just \$ 12,500 . along with teaching law , mcelroy is also an accomplished author with a number of published biographies and children 's books . has been placed on leave while school investigates .} 
& \multicolumn{1}{c|}{2.3270} 
& - \\ \midrule

\multicolumn{1}{c|}{\textbf{\begin{tabular}[c]{@{}c@{}}Self-generation\\ (BART-large+cnn)\end{tabular}}} & \multicolumn{1}{m{13.2cm}|}{lisa mcelroy, a 50-year-old drexel professor, appeared on the show in 2010 while it was still hosted my meredith vieira. she's accused of sending the inappropriate message this past march 31 under the subject line: 'great article on writing briefs' when recipients opened the enclosed link, philly.com reports that they were directed to a video of 'a woman engaging in a sexually explicit act' the married mother-of-two has been placed on administrative leave.} 
& \multicolumn{1}{c|}{5.1158} 
& +2.7888 \\ \midrule

\multicolumn{1}{c|}{\textbf{\begin{tabular}[c]{@{}c@{}}BERTSum\\(Factuality=1.0)\end{tabular}}} & \multicolumn{1}{m{13.2cm}|}{lisa mcelroy , 50 , who teaches legal writing at drexel university , appeared on the show in 2010 while it was still hosted my meredith vieira . she got tripped up on the \$ 12,500 level after flying through the first few questions , philly.com reports . mcelroy answered wrong and walked away with around \$ 5,000 .} 
& \multicolumn{1}{c|}{4.5153} 
& +2.1883  \\ \midrule

\multicolumn{1}{c|}{\textbf{\begin{tabular}[c]{@{}c@{}}Reference\\ (w/ wrong subject)\end{tabular}}} & \multicolumn{1}{m{13.2cm}|}{lisa mcelroy , 50 , who teaches legal writing at \textcolor{red}{philadelphia} university , reportedly sent the ` inappropriate ' message on march 31 . when recipients clicked the enclosed link , they were allegedly directed to a video of ' a woman engaging in a sexually explicit act ' . mcelroy appeared on the popular game show in 2010 with then-host meredith vieira but lost the game after reaching just \$ 12,500 . along with teaching law , mcelroy is also an accomplished author with a number of published biographies and children 's books . has been placed on leave while school investigates .} 
& \multicolumn{1}{c|}{2.2040} 
& -0.1230 \\ \midrule

\multicolumn{1}{c|}{\textbf{\begin{tabular}[c]{@{}c@{}}Reference\\ (w/ negation)\end{tabular}}} & \multicolumn{1}{m{13.2cm}|}{lisa mcelroy , 50 , who teaches legal writing at drexel university , reportedly \textcolor{red}{did not} send the ` inappropriate ' message on march 31 . when recipients clicked the enclosed link , they were allegedly directed to a video of ' a woman engaging in a sexually explicit act ' . mcelroy appeared on the popular game show in 2010 with then-host meredith vieira but lost the game after reaching just \$ 12,500 . along with teaching law , mcelroy is also an accomplished author with a number of published biographies and children 's books . has been placed on leave while school investigates .} 
& \multicolumn{1}{c|}{2.0178} 
& -0.3092 \\ \bottomrule

\end{tabu}}
\caption{\label{tab:qual_pos2}Testing $\mathrm{HaRiM}^+$ metric under hallucination detecting scenario. Part of the source article irrelevant to the summaries are omitted for clarity. The words highlighted red are hallucinated information deliberately injected to the reference. BERTSum refers to extractive summarization system suggested in \cite{liu-lapata-2019-text}.}
\end{table*}

\end{document}